\documentclass{article}

    \PassOptionsToPackage{numbers, compress}{natbib}
\usepackage{hyperref}
\usepackage{amsmath}
\usepackage{amssymb}
\usepackage{cleveref}

\usepackage{multirow, graphicx, rotating}
\usepackage{adjustbox}
\usepackage{multirow}

\usepackage{pifont}
\usepackage{graphicx}
\usepackage{subcaption}
\usepackage{placeins}
\usepackage{array}
\usepackage{xcolor}
\usepackage{pifont}
\usepackage{amssymb}

\newcommand{\cmark}{\textcolor{green!60!black}{\checkmark}}
\newcommand{\xmark}{\textcolor{red!75!black}{\ding{55}}}

 \usepackage[preprint]{neurips_2026}

\usepackage[utf8]{inputenc} % allow utf-8 input
\usepackage[T1]{fontenc}    % use 8-bit T1 fonts
\usepackage{hyperref}       % hyperlinks
\usepackage{url}            % simple URL typesetting
\usepackage{booktabs}       % professional-quality tables
\usepackage{amsfonts}       % blackboard math symbols
\usepackage{nicefrac}       % compact symbols for 1/2, etc.
\usepackage{microtype}      % microtypography
\usepackage{xcolor}         % colors

\title{Importance-Aware Pruning for Diffusion Models}

\author{
Ba-Thinh Lam \quad \quad \quad
Srijan Das \quad \quad \quad
Hieu Le \\
Department of Machine Learning, UNC Charlotte, NC, USA \\
\texttt{\{blam5,sdas24,hle40\}@charlotte.edu} }

\begin{document}

% !TEX root = ../top.tex
% !TEX spellcheck = en-US

\newif\ifdraft
\draftfalse
\drafttrue % <---- The way to switch between draft mode is to comment this

\definecolor{orange}{rgb}{1,0.5,0}
\definecolor{violet}{RGB}{70,0,170}
\definecolor{magenta}{RGB}{170,0,170}
\definecolor{dgreen}{RGB}{0,150,0}

\ifdraft
 \newcommand{\PF}[1]{{\color{red}{\bf PF: #1}}}
 \newcommand{\pf}[1]{{\color{red} #1}}
 \newcommand{\FS}[1]{{\color{blue}{\bf FS: #1}}}
 \newcommand{\fs}[1]{{\color{blue} #1}}
 \newcommand{\hl}[1]{{\color{orange} #1}}
 \newcommand{\HL}[1]{{\color{orange}{\bf HL: #1}}}
 \newcommand{\BG}[1]{{\color{olive}{\bf BG: #1}}}
 \newcommand{\bg}[1]{{\color{olive} #1}}
 \newcommand{\red}[1]{{\color{red}#1}}
 \newcommand{\todo}[1]{{\color{red}#1}}
 \newcommand{\TODO}[1]{\textbf{\color{red}[TODO: #1]}}
 \newcommand{\NT}[1]{{\color{violet}{\bf NT: #1}}}
 \newcommand{\nt}[1]{{\color{violet} #1}}
\else
 \newcommand{\PF}[1]{}
 \newcommand{\pf}[1]{#1}
 \newcommand{\FS}[1]{}
 \newcommand{\fs}[1]{#1}
 \newcommand{\hl}[1]{#1}
  \newcommand{\HL}[1]{}
 \newcommand{\BG}[1]{}
 \newcommand{\bg}[1]{#1}
 \newcommand{\ME}[1]{}
  \newcommand{\me}[1]{#1}
  \newcommand{\TODO}[1]{}
  \newcommand{\todo}[1]{#1}
  \newcommand{\NT}[1]{{\color{violet}{}}}
  \newcommand{\nt}[1]{ #1 }
\fi

\newcommand{\comment}[1]{}
\newcommand{\parag}[1]{\vspace{-3mm}\paragraph{#1}}
\newcommand{\sparag}[1]{\vspace{-3mm}\subparagraph{#1}}
\renewcommand{\floatpagefraction}{.99}

% Matrices
\newcommand{\bA}{\mathbf{A}}
\newcommand{\bC}{\mathbf{C}}
\newcommand{\bD}{\mathbf{D}}
\newcommand{\bH}{\mathbf{H}}
\newcommand{\bK}{\mathbf{K}}
\newcommand{\bP}{\mathbf{P}}
\newcommand{\bR}{\mathbf{R}}
\newcommand{\bX}{\mathbf{X}}
\newcommand{\bZ}{\mathbf{Z}}

\newcommand{\real}{\mathbb{R}}

\newcommand{\bc}{\mathbf{c}}
\newcommand{\f}{\mathbf{f}}
\newcommand{\bI}{\mathbf{I}}
\newcommand{\bm}{\mathbf{m}}
\newcommand{\bs}{\mathbf{s}}
\newcommand{\bt}{\mathbf{t}}
\newcommand{\bu}{\mathbf{u}}
\newcommand{\bw}{\mathbf{w}}
\newcommand{\bx}{\mathbf{x}}
\newcommand{\by}{\mathbf{y}}
\newcommand{\bz}{\mathbf{z}}

\newcommand{\radius}{\mathbf{r}}

\newcommand{\cF}{\mathcal F}
\newcommand{\fd}{\mathcal{F}_{d}}
\newcommand{\fz}{\mathcal{F}_{z}}

\newcommand{\OURS}[0]{\textbf{OURS}}
\newcommand{\FGSMU}[1]{\textbf{FGSM-U(#1)}}
\newcommand{\FGSMT}[1]{\textbf{FGSM-T(#1)}}
\newcommand{\FGSMUE}[1]{\textbf{FGSM-UE(#1)}}
\newcommand{\FGSMTE}[1]{\textbf{FGSM-TE(#1)}}

% Vectors:
\newcommand{\colvecTwo}[2]{\ensuremath{
		\begin{bmatrix}{#1}	\\	{#2}	\end{bmatrix}
}}
\newcommand{\colvec}[3]{\ensuremath{
		\begin{bmatrix}{#1}	\\	{#2}	\\	{#3} \end{bmatrix}
}}
\newcommand{\colvecFour}[4]{\ensuremath{
		\begin{bmatrix}{#1}	\\	{#2}	\\	{#3} \\	{#4}	\end{bmatrix}
}}

\newcommand{\rowvecTwo}[2]{\ensuremath{
		\begin{bmatrix}{#1}	&	{#2}	\end{bmatrix}
}}
\newcommand{\rowvec}[3]{\ensuremath{
		\begin{bmatrix}{#1} &	{#2}	&	{#3} \end{bmatrix}
}}
\newcommand{\rowvecFour}[4]{\ensuremath{
		\begin{bmatrix}{#1}	&	{#2}	&	{#3} &	{#4}	\end{bmatrix}
}}

% Transpose:
\newcommand{\tr}{^\intercal}

% Temporary, to store default tabcolsep
\newlength{\mytabcolsep}
\setlength\mytabcolsep{\tabcolsep}

\newcommand{\visStructured}[2]{
\begin{figure*}[!ht]
    \centering
    \small
    
    % Define image width (3 columns * 0.31 = ~0.93\textwidth, leaving space for labels)
    \def\imgw{0.31\textwidth}

    % 1 column for labels, 3 columns for images
    \begin{tabular}{@{} >{\centering\arraybackslash}m{15pt} c c c @{}}
        
        % --- Headers ---
        & \textbf{20\% Sparsity} & \textbf{30\% Sparsity} & \textbf{40\% Sparsity} \\[1ex]

        % --- Row 1: OBS-Diff ---
        \rotatebox{90}{\textbf{OBS-Diff}} &
        \adjustbox{valign=m}{\includegraphics[width=\imgw]{figures/Structured_pruning/#1/OBS-Diff_20pct.jpg}} &
        \adjustbox{valign=m}{\includegraphics[width=\imgw]{figures/Structured_pruning/#1/OBS-Diff_30pct.jpg}} &
        \adjustbox{valign=m}{\includegraphics[width=\imgw]{figures/Structured_pruning/#1/OBS-Diff_40pct.jpg}} \\[1.5ex] % Gap between rows

        % --- Row 2: Ours (CFG) ---
        \rotatebox{90}{\textbf{Ours}} &
        \adjustbox{valign=m}{\includegraphics[width=\imgw]{figures/Structured_pruning/#1/Ours_CFG_20pct.jpg}} &
        \adjustbox{valign=m}{\includegraphics[width=\imgw]{figures/Structured_pruning/#1/Ours_CFG_30pct.jpg}} &
        \adjustbox{valign=m}{\includegraphics[width=\imgw]{figures/Structured_pruning/#1/Ours_CFG_40pct.jpg}} \\
        
    \end{tabular}

    \vspace{2mm}
    \caption{\textbf{Qualitative resutls on PixArt-$\Sigma$.}
Comparison between OBS-Diff and our importance-aware pruning guided by CFG \cite{ho2022classifier} signals at 20\%, 30\%, and 40\% structured sparsity. \textbf{Prompt:} "#2"}
    \label{fig:structured_#1}
\end{figure*}
}

\maketitle

\newcommand{\TL}[1]{{\color{green}{\bf TL: #1}}} 

\begin{figure*}[!ht]
    \centering
    \small
    
    % Define widths
    \def\bigimgw{0.30\textwidth}
    \def\smallimgw{0.15\textwidth}

    % Outer table
    \begin{tabular}{@{} c c @{\hspace{15pt}} c @{}}
        
        % --- Headers ---
        {Dense} & {Important Mask (CFG)} & 
        \begin{tabular}{@{} p{\smallimgw} p{\smallimgw} @{\hspace{8pt}} c @{}}
            \centering {30\% Sparsity} & \centering {40\% Sparsity} & 
        \end{tabular} \\ [1ex]

        % --- Images ---
        % Left: Large Dense Image
        \adjustbox{valign=m}{\includegraphics[width=\bigimgw]{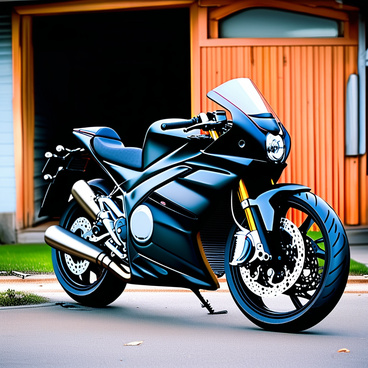}} & 
        
        % Middle: Large Signal Image
        \adjustbox{valign=m}{\includegraphics[width=\bigimgw]{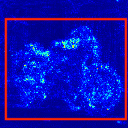}} & 
        
        % Right: The 2x2 Grid packed into a single centered cell
        \adjustbox{valign=m}{%
            \begin{tabular}{@{} c c @{\hspace{5pt}} c @{}}
                % Row 1 (OBS-Diff) - Everything vertically centered!
                \adjustbox{valign=m}{\includegraphics[width=\smallimgw]{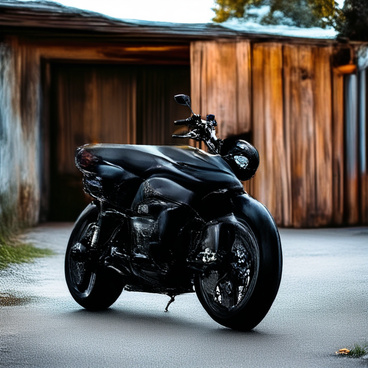}} & 
                \adjustbox{valign=m}{\includegraphics[width=\smallimgw]{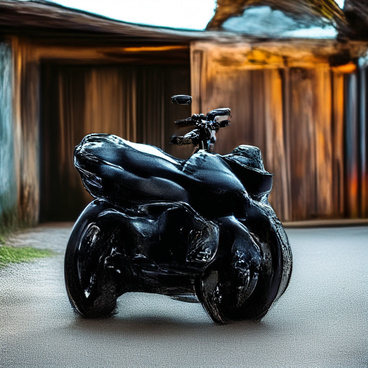}} & 
                \adjustbox{valign=m}{\rotatebox[origin=c]{270}{{OBS-Diff\cite{zhu2025obs}}}} \\ [2pt] % <-- TIGHT GAP HERE
                
                % Row 2 (Ours) - Everything vertically centered!
                \adjustbox{valign=m}{\includegraphics[width=\smallimgw]{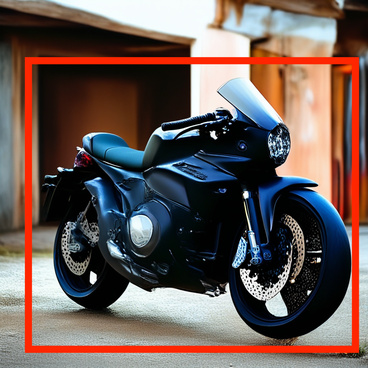}} & 
                \adjustbox{valign=m}{\includegraphics[width=\smallimgw]{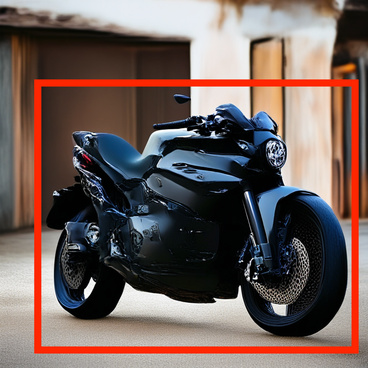}} & 
                \adjustbox{valign=m}{\rotatebox[origin=c]{270}{{Ours}}} \\
            \end{tabular}%
        } \\
    \end{tabular}\\
    %\vspace{3mm}
    %\textit{\textbf{Prompt:} ``A black Honda motorcycle parked in front of a garage".}
    \caption{\textbf{Qualitative comparison on PixArt-$\Sigma$ for Structured Pruning.} Our method (bottom) preserves subject integrity substantially better than the OBS-Diff baseline (top) by retaining semantically important regions, highlighted by red boxes. We achieve this by injecting importance signals such as classifier-free guidance (CFG)~\cite{ho2022classifier} into the pruning objective.}
    \label{fig:teaser}
\end{figure*}

\begin{abstract} 

We propose importance-aware pruning for diffusion models, a training-free framework that prioritizes preserving parameters critical to semantically salient image regions. To do so, we incorporate spatial importance maps- derived from conditioning signals or model attention- into the pruning objective. This produces parameter rankings aligned with perceptual relevance rather than uniform reconstruction error. On the MS-COCO dataset, our proposed approach consistently retains subject fidelity and structural correctness at high compression ratios where conventional pruning causes visible degradation. These results demonstrate that content-aware objectives are key to perceptually faithful compression of generative models.
  % \keywords{First keyword \and Second keyword \and Third keyword}
\end{abstract}

\section{Introduction}

Pruning diffusion models~\cite{dhariwal2021diffusion,chen2023pixart,esser2024scaling,li2024hunyuan,li2024efficient,tang2025exploring,xie2025sana,labs2025flux1kontextflowmatching} is an important problem since  diffusion models contain billions of parameters and require dozens of denoising steps at inference, making them expensive for many applications. Model pruning offers a promising solution by removing redundant parameters while preserving generation quality. Recent methods, such as OBS-Diff~\cite{zhu2025obs}, have demonstrated that large-scale diffusion models can be pruned effectively without retraining.

Existing pruning methods typically treat all spatial locations equally. Yet not all regions of a generated image contribute equally to its perceived quality. Errors in prompt-relevant objects, faces, or foreground structures can significantly affect the perceived quality and semantic correctness of an image, while similar errors in background regions may be less noticeable~\cite{Itti1998AMO,Xue2025FewshotPS}. This distinction becomes particularly important at high sparsity levels, where limited model capacity makes it difficult to preserve all visual details. As illustrated in Fig.~\ref{fig:teaser}, conventional pruning can preserve the overall appearance of an image while degrading important subject details. This raises a fundamental question: Can model pruning prioritize the content that matters most for generation?

In this work, we propose importance-aware pruning, a training-free framework that incorporates spatial importance into parameter selection. To identify these regions, we construct spatial importance maps using signals such as classifier-free guidance (CFG)~\cite{ho2022classifier}, Canny edge detection~\cite{Canny1986ACA}, or object detection~\cite{10533619}. For example, CFG highlights regions strongly influenced by the text prompt, as illustrated in Fig.~\ref{fig:teaser}. These signals allow the pruning process to focus on different aspects of image content depending on the desired preservation goal. We then incorporate these importance maps into the pruning objective by assigning greater weight to reconstruction errors in important regions. For linear layers, this can be implemented by weighting the input activations used to estimate parameter importance. The framework therefore integrates naturally with second-order pruning methods such as OBS-Diff while retaining their parameter-selection and weight-compensation procedures. The resulting method remains one-shot and requires no retraining.

 Our results show consistent improvements over previous methods, particularly at high sparsity levels, where our method better preserves subject integrity and structural details. We further demonstrate that different importance signals can prioritize different visual properties. In addition, category-targeted calibration allows pruning to favor specific semantic concepts, such as cats, airplanes, or people. More broadly, our work challenges the long-standing assumption that pruning should minimize average loss and instead argues that content-aware objectives are essential for perceptually faithful compression, connecting low-level parameter pruning with high-level semantic importance.

Contributions:
\begin{itemize}
\item We propose importance-aware pruning for diffusion models, grounded in the principle that parameters responsible for semantically critical image regions should be preferentially preserved during compression.
\item We introduce a simple yet effective framework that supplements OBS saliency with spatial importance signals, and show that the signals can be flexibly derived from a variety of sources depending on the desired preservation goal.
\item We show consistent improvements over the state-of-the-art method, OBS-Diff, across diffusion backbones and sparsity levels, with especially clear gains in subject fidelity and structural preservation under high compression ratios, for both structured and unstructured pruning.

\end{itemize}
\section{Related Work}
\label{sec:related_work}

\paragraph{Diffusion Models.}
Diffusion models~\cite{sohl2015deep, dhariwal2021diffusion, ho2020denoising, song2020denoising} generate data by progressively transforming noise into structured samples through a learned reverse diffusion process. During training, clean data are gradually corrupted and the model learns to denoise across timesteps. This formulation has enabled state-of-the-art results across diverse generative tasks, including image synthesis~\cite{rombach2022high, saharia2022photorealistic, peebles2023scalable, chen2023pixartalpha, Ramesh2022HierarchicalTI, nichol2021glide, graikos2024learned, meng2021sdedit, xue2024raphael, balaji2022ediff, karras2025guiding,Xu2024ASQ}, video generation~\cite{singer2022make, blattmann2023stable, guo2023animatediff, opensora, pku_yuan_lab_and_tuzhan_ai_etc_2024_10948109, brooks2024video}, 3D synthesis~\cite{poole2022dreamfusion, liu2023zero, long2024wonder3d, shi2023zero123++, li20223ddesigner, luo2021diffusion, nichol2022point}, and audio modeling~\cite{kong2020diffwave, huang2023make}. 
Many practical systems follow the latent diffusion design~\cite{rombach2022high}, where images are encoded into latent tokens processed by a U-Net backbone augmented with attention and transformer layers. Extensions adapt this structure for multi-view~\cite{shi2023zero123++} and video generation~\cite{guo2023animatediff}, while recent work replaces convolutional backbones with fully transformer-based diffusion architectures for improved scalability~\cite{chen2023pixartalpha, opensora}. Classifier-free guidance (CFG)~\cite{ho2022classifier} improves conditional fidelity by combining conditional and unconditional predictions during sampling, and prior work suggests CFG implicitly highlights semantically important regions~\cite{zhao2023magicfusion, wang2024high}. We build on this observation but use importance signals to guide parameter pruning rather than generation.

\paragraph{Diffusion Acceleration.}
Due to the high computational cost of diffusion inference, a wide range of efficiency strategies have been explored. Some methods require retraining, including improved architectures~\cite{rombach2022high, pernias2023wurstchen, kim2023architectural}, structured pruning~\cite{fang2023structural}, compression approaches~\cite{zhao2023mobilediffusion, yang2023diffusion, li2024snapfusion}, step distillation~\cite{salimans2022progressive, meng2023distillation, sauer2025adversarial, liu2023instaflow, habibian2024clockwork}, and consistency modeling~\cite{song2023consistency, luo2023latent}. Other approaches improve efficiency without retraining, such as advanced samplers~\cite{song2020denoising, liu2022pseudo, lu2022dpm, lu2022dpm_pp}, feature caching~\cite{ma2024deepcache, wimbauer2024cache, chen2024delta, zhao2024real, kahatapitiya2024adaptive, so2023frdiff, lv2024fastercache, li2023faster}, quantization~\cite{li2023q, chen2024q, he2024ptqd, wang2024quest, deng2024vq4dit, so2024temporal}, mixed-resolution inference~\cite{wu2026phase}, and token reduction~\cite{kahatapitiya2025object, Bolya2023TokenMF, Bolya2022TokenMY}. In particular, recent studies adapt token reduction to diffusion inference~\cite{Bolya2023TokenMF, kim2024token, haurum2025agglomerative, wang2024attention, li2024vidtome, kahatapitiya2025object, zhao2024dynamic, liang2024looking, wang2024sparsedm, smith2024todo, lutoma}, reducing computation by pruning or merging intermediate tokens during sampling. While effective for accelerating inference, these approaches operate on intermediate representations rather than model parameters and therefore do not reduce model size or memory footprint. Moreover, token selection is often heuristic or spatially uniform, which can degrade fidelity in visually important regions. In contrast, we study parameter-level pruning and explicitly bias pruning toward preserving semantically salient image content. Closest to our work is the importance-based token merging method of Wu \textit{et al.}~\cite{Wu_2025_ICCV}, which prioritizes less important tokens for merging based on CFG guidance. However, token merging is orthogonal to model pruning and, in practice, token merging can be applied on top of a pruned model for further acceleration.

\paragraph{Model Pruning.}
Pruning removes redundant parameters to reduce computation while maintaining model performance. Early methods rely on magnitude-based criteria~\citep{han2015learning} or first-order approximations, whereas second-order approaches such as Optimal Brain Damage~\citep{lecun1989optimal} and Optimal Brain Surgeon~\citep{hassibi1992second} estimate parameter saliency using curvature information to better predict the effect of removal. Layer-wise formulations~\citep{dong2017learning, frantar2022optimal} made second-order pruning tractable at scale, and subsequent methods such as SparseGPT~\citep{frantar2023sparsegpt} and Wanda~\citep{sun2024wanda} extended these ideas to billion-parameter language models. For diffusion models, existing approaches are either tailored to specific architectures~\citep{fang2023structural, li2023snapfusion, kim2024bksdm, castells2024ld, zhang2024laptop, zhao2023mobilediffusion}, dependent on costly retraining~\citep{fang2023structural, zhu2024dip, fang2025tinyfusion, zhang2024ecodiff, zhao2023mobilediffusion}, or limited to structured sparsity~\cite{kim2024bksdm, zhang2024laptop, zhao2023mobilediffusion, ganjdanesh2024not, zhu2024dip, chen2025adversarial}. OBS-Diff~\citep{zhu2025obs} is the first to apply second-order pruning to large-scale text-to-image diffusion models in a one-shot, training-free manner. However, its pruning objective treats all spatial locations uniformly, without regard for perceptual relevance. Our work extends this line by incorporating spatial importance signals into the saliency estimation, aligning pruning decisions with what matters most in the generated output.
% \HL{need to add more diffusion pruning methods - borrowing from obs diff} \TL{Okay} \TL{Edited, do you think should add more?}/HL{just make sure youve added all diffusion ones}

% \paragraph{Importance-Aware Computation.}
% A related direction adapts computation based on input saliency, often by prioritizing informative regions or tokens. Token reduction methods~\cite{haurum2023tokens, Bolya2022TokenMY, wang2024zero} reduce computation by pruning or merging tokens and have been explored across classification~\cite{marin2023token, Bolya2022TokenMY, haurum2023tokens, liang2022not}, segmentation~\cite{xu2022groupvit, kienzle2024segformer++}, detection~\cite{liu2024revisiting}, video understanding~\cite{choi2024vid}, and language models~\cite{li2024tokenpacker, jin2024chat, song2024less}. Methods for identifying unimportant tokens include attention-based scoring~\cite{fayyaz2022adaptive, haurum2023tokens, liang2022not, long2023beyond, wei2023joint, wu2023ppt, xu2022evo}, differentiable selection~\cite{kong2022spvit,  rao2021dynamicvit, wei2023joint}, sampling strategies~\cite{fayyaz2022adaptive, yin2022vit}, sparsity regularization~\cite{chen2021chasing, li2022sait}, and reinforcement learning approaches~\cite{pan2023interpretability}. These works demonstrate that importance signals can guide efficient computation, but they primarily operate at the feature or token level. In contrast, our approach leverages importance to guide which model parameters are removed.

\section{Method}
\label{sec:method}

We propose importance-aware pruning, a framework that incorporates spatial importance signals into the OBS reconstruction objective. By weighting reconstruction errors according to spatial importance, the framework biases parameter selection toward preserving semantically or structurally important regions.

%------------------------------------------------
%\subsection{Preliminaries}

\paragraph{Preliminaries.} 
For a layer $l$ with weight matrix $W_l$ and input activations $X_{l,t}$ at 
timestep $t$, the layer-wise pruning objective seeks a sparse weight matrix 
$\hat{W}_l$ that minimizes the reconstruction error:
\begin{equation}
\hat{W}_l^* = \arg\min_{\hat{W}_l} \;
\mathbb{E}_{t \sim [1,T]}\left[\alpha_t
\|\hat{W}_l X_{l,t} - W_l X_{l,t}\|_2^2\right]
\quad \text{s.t.} \quad \text{Sparsity}(\hat{W}_l) = S,
\end{equation}
where $\alpha_t$ is a timestep-dependent weight that places greater importance 
on earlier denoising steps, following the logarithmically decreasing schedule 
of \cite{zhu2025obs}. 
\subsection{OBS Pruning}
The Optimal Brain Surgeon (OBS) framework solves the pruning objective by 
approximating the change in loss incurred by removing weight $w_q$ via a 
second-order Taylor expansion, yielding a closed-form saliency score:
\begin{equation}
S_{l,q} = \frac{w_q^2}{2[H_l^{-1}]_{qq}}.
\end{equation}
Parameters with smaller saliency 
are pruned first, with the remaining weights updated analytically to compensate 
for each removal via efficient Cholesky-based inverse Hessian updates 
\citep{frantar2023sparsegpt}.

However, this saliency estimate assumes that the model has converged to a minimum with negligible first-order gradient, which is rarely true in practice for diffusion models. As a result, the saliency scores are only approximate and may not faithfully reflect the true cost of removing each parameter. This motivates incorporating external importance guidance into the saliency estimation process. Specifically, we reweight the Hessian construction with a spatial importance signal, biasing saliency toward parameters that affect perceptually critical regions. This produces a pruning criterion that is better aligned with what the model should preserve, while keeping the structure and tractability of the OBS pipeline unchanged.

\subsection{Spatial Importance Signal}

Let $M_t \in \mathbb{R}^{H \times W}$ denote a spatial importance map at timestep $t$, with each entry reflecting the perceptual relevance of the corresponding spatial location. We construct this map from the classifier-free guidance (CFG) difference \citep{ho2022classifier}:
\begin{equation}
\Delta_t =
\epsilon_\theta(\mathbf{x}_t,t,y)
-
\epsilon_\theta(\mathbf{x}_t,t,\varnothing),
\end{equation}
where $y$ denotes the text condition and $\varnothing$ denotes the unconditional input. The magnitude of this difference indicates how strongly the prompt influences the prediction at each location and serves as a proxy for spatial importance.

The raw CFG difference retains the latent channel dimension. We obtain a two-dimensional spatial map by averaging its absolute values across the $C$ channels:
\begin{equation}
\tilde{M}_t(i,j) =
\frac{1}{C}\sum_{k=1}^{C}
\left|\Delta_{t,k,i,j}\right|.
\end{equation}
We then normalize the map independently for each sample and timestep:
\begin{equation}
M_t =
\frac{\tilde{M}_t - \min(\tilde{M}_t)}
{\max(\tilde{M}_t) - \min(\tilde{M}_t) + \epsilon},
\end{equation}
where the minimum and maximum are taken over spatial locations and $\epsilon > 0$ ensures numerical stability.

Our framework is not restricted to CFG-based maps. Other signals that encode spatial relevance, such as Canny edge maps or object detection masks, can be used depending on the desired preservation goal, as demonstrated in Section~\ref{sec:exp}.

% %------------------------------------------------
% \subsection{Spatial Importance Signal}

% Let $M_t \in \mathbb{R}^{H \times W}$ denote a spatial importance map at 
% timestep $t$, where each entry reflects the perceptual relevance of the 
% corresponding spatial location. We compute $M_t$ as the magnitude of the 
% classifier-free guidance (CFG) delta \cite{ho2022classifier}:
% \begin{equation}
% M_t = \left|\epsilon_\theta(\mathbf{x}_t, t, c) - 
% \epsilon_\theta(\mathbf{x}_t, t)\right|,
% \end{equation}
% which directly encodes where the prompt influences the output at each denoising 
% step, concentrated over semantically salient regions and diminished elsewhere. 
% We note that our framework is not restricted to this choice: $M_t$ can be derived from any 
% signal that encodes spatial relevance, such as Canny edge maps or object 
% detection masks, depending on the desired preservation goal- as we will show in Sec. \ref{sec:exp}.

% The raw CFG difference retains the latent channel dimension. We convert it into a final 2D spatial map by averaging the absolute guidance response over channels:
% \begin{equation}
% \tilde{M}_t(i,j) =
% \frac{1}{C}\sum_{c=1}^{C}
% \left|
% \epsilon_\theta(x_t,t,y)_{c,i,j}
% -
% \epsilon_\theta(x_t,t,\varnothing)_{c,i,j}
% \right|.
% \end{equation}
% We then normalize the map independently for each sample and timestep:
% \begin{equation}
% M_t =
% \frac{\tilde{M}_t - \min(\tilde{M}_t)}
% {\max(\tilde{M}_t)-\min(\tilde{M}_t)+\epsilon}.
% \end{equation}

\subsection{Importance-Aware Hessian Construction}

We incorporate the spatial importance signal into the pruning objective by 
weighting the reconstruction error at each spatial location by its importance:
\begin{equation}
\hat{W}_l^* = \arg\min_{\hat{W}_l} \;
\sum_{t=1}^{T} \alpha_t
\left\|M_t \odot \left(\hat{W}_l X_{l,t} - W_l X_{l,t}\right)\right\|_2^2
\quad \text{s.t.} \quad \text{Sparsity}(\hat{W}_l) = S.
\end{equation}
For linear layers, since $W_l$ acts along the channel dimension independently 
of the spatial dimension, it commutes with $M_t$, allowing us to define 
importance-filtered activations $X'_{l,t} = M_t \odot X_{l,t}$ and rewrite 
the objective as a standard OBS problem with $X_{l,t}$ replaced by $X'_{l,t}$. 
The importance-aware Hessian follows directly (see Appendix \ref{app:derivation} for 
the full derivation):
\begin{equation}
H_{l,\textit{imp}} = 2\sum_{t=1}^{T} \alpha_t 
\mathbb{E}\left[X'_{l,t} X_{l,t}^{\prime\top}\right],
\end{equation}
and substituting into the OBS saliency formula gives:
\begin{equation}
S_{l,q} = \frac{w_q^2}{2[H_{l,\textit{imp}}^{-1}]_{qq}}.
\end{equation}
Parameters whose activations concentrate in high-importance regions accumulate 
larger curvature and are preferentially retained; those whose influence is 
concentrated in perceptually irrelevant areas are preferentially removed. The 
entire modification reduces to a single substitution in the Hessian computation, 
with no change to the pruning pipeline or its computational complexity.

\section{Experiments}
\label{sec:exp}
%\vspace{-5mm}
\subsection{Experimental Settings}
\label{ssec:exp_setting}

\parag{Model and Dataset.} We evaluate our method using {Stable Diffusion 3 (SD3) Medium} \cite{esser2024scaling} and {PixArt-$\Sigma$} \cite{chen2024pixart} as the base generative models. For the pruning calibration phase, we utilize 1,000 samples from the {GCC3M} dataset \cite{sharma2018conceptual}. To evaluate generative performance, we generate images using 1,000 prompts from the {MS-COCO 2017} \cite{lin2014microsoft} validation set. 

\parag{Baselines and Metrics.}
We compare our importance-aware framework against OBS-Diff~\cite{zhu2025obs}, the current state-of-the-art second-order pruning method for diffusion models based on the Optimal Brain Surgeon framework~\cite{hassibi1992second}. To evaluate generated image quality, we report CLIP Score~\cite{hessel2021clipscore} as a proxy for text alignment and ImageReward~\cite{xu2023imagereward} as a human-aligned perceptual quality metric. For the category-targeted pruning experiment, we further assess semantic visual quality using MUSIQ~\cite{ke2021musiq}.

\parag{Hyper-parameter setting.} During the pruning phase, we set the classifier-free guidance (CFG) scale to $7.0$ and $4.5$ for SD3-Medium and PixArt-$\Sigma$, respectively, while setting the CFG scale to $7.0$ for both during testing. We use $10$ and $25$ denoising steps for the pruning and testing phases, respectively, across all experiments. We choose a batch size of $2$ during the pruning phase.

\subsection{Quantitative Results}
\vspace{-5mm}

\begin{table}[!htp]
\centering
\caption{Quantitative results of unstructured pruning of SD3-Medium on the GCC3M dataset~\cite{sharma2018conceptual}.}
\label{tab:1000_SD3_final}
\setlength{\tabcolsep}{0pt}
\begin{tabular*}{\textwidth}{@{\extracolsep{\fill}} lllccc}
\toprule
Model & Sparsity & Method & CLIP Score $\uparrow$ & ImageReward $\uparrow$ & MUSIQ $\uparrow$ \\ 
\midrule

\multirow{9}{*}{SD3-Medium} 
& \multicolumn{2}{l}{Dense Model} & 32.26 & 0.98 & 72.68 \\ 
\cmidrule{2-6} 

& \multirow{2}{*}{30\%} 
& OBS-Diff & \textbf{32.25} & 0.96 & \textbf{72.53} \\
&  
& Ours (CFG) & 32.24 & \textbf{0.98} & 72.41 \\ 
\cmidrule{2-6} 

& \multirow{2}{*}{40\%} 
& OBS-Diff & 32.30 & 0.92 & 71.02 \\
&  
& Ours (CFG) & \textbf{32.35} & \textbf{0.95} & \textbf{71.48} \\ 
\cmidrule{2-6} 

& \multirow{2}{*}{45\%} 
& OBS-Diff & 32.25 & 0.84 & 68.76 \\
&  
& Ours (CFG) & \textbf{32.30} & \textbf{0.87} & \textbf{69.86} \\ 
\cmidrule{2-6} 

& \multirow{2}{*}{50\%} 
& OBS-Diff & 32.16 & 0.71 & 65.98 \\
&  
& Ours (CFG) & \textbf{32.20} & \textbf{0.76} & \textbf{66.30} \\ 
\bottomrule
\end{tabular*}
\end{table}

\begin{figure*}[!htp]
    \centering
    \small

    % Stable width control
    \newlength{\densew}
    \newlength{\denseimgw}
    \newlength{\imgw}
    \setlength{\densew}{0.36\textwidth}      % dense column width
    \setlength{\denseimgw}{0.30\textwidth}   % actual dense image size
    \setlength{\imgw}{0.17\textwidth}        % each pruned image

    % ================= Dense column =================
    \begin{minipage}[c]{\densew}
        \centering
        {Dense}\\[0.8ex]
        \includegraphics[width=\denseimgw]{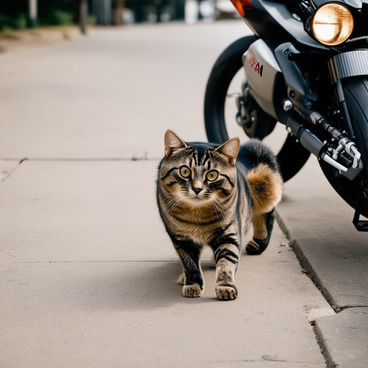}

        \vspace{1mm}
        {\footnotesize
        \textbf{Prompt:} `` A cat walks next to a motorcycle on a sidewalk.''}
    \end{minipage}
    \hspace{0.015\textwidth}
    % ================= Pruned grid =================
    \begin{minipage}[c]{0.60\textwidth}
        \centering

        % Row 1: OBS-Diff
        \includegraphics[width=\imgw]{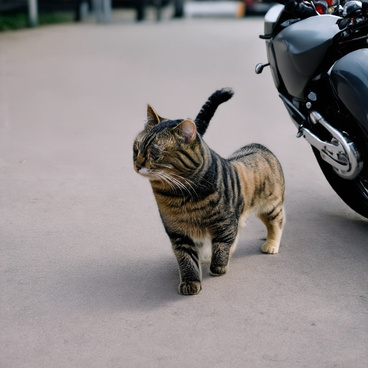}
        \hspace{3pt}
        \includegraphics[width=\imgw]{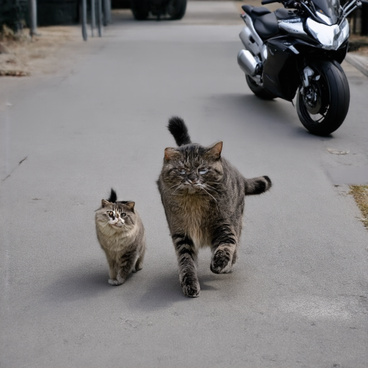}
        \hspace{3pt}
        \includegraphics[width=\imgw]{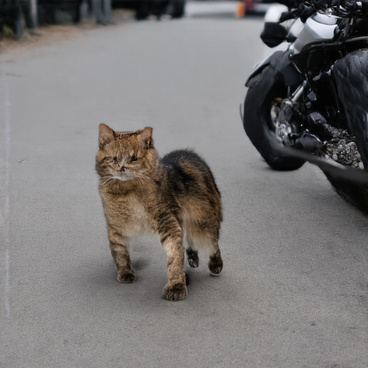}
        \hspace{2pt}
        \raisebox{0.45\imgw}{\rotatebox[origin=c]{270}{OBS-Diff~\cite{zhu2025obs}}}
        \\[2pt]

        % Row 2: Ours
        \includegraphics[width=\imgw]{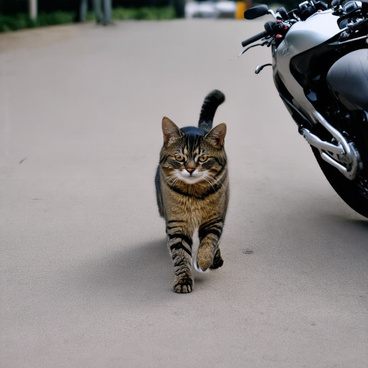}
        \hspace{3pt}
        \includegraphics[width=\imgw]{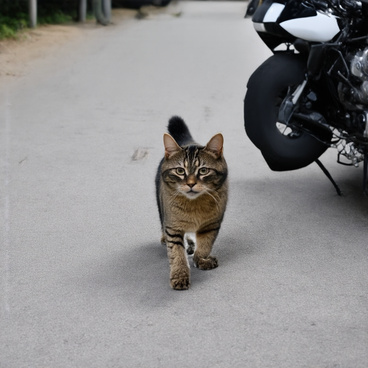}
        \hspace{3pt}
        \includegraphics[width=\imgw]{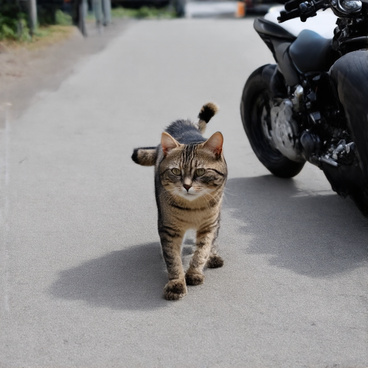}
        \hspace{2pt}
        \raisebox{0.45\imgw}{\rotatebox[origin=c]{270}{{Ours}}}
        % Column headers
        \makebox[\imgw][c]{40\% Sparsity}
        \hspace{3pt}
        \makebox[\imgw][c]{45\% Sparsity}
        \hspace{3pt}
        \makebox[\imgw][c]{50\% Sparsity}
        \hspace{0.025\textwidth}
        \\[0.8ex]

    \end{minipage}
    
    \caption{\textbf{Qualitative comparison across different sparsity levels on SD3-Medium.}
    The top row shows OBS-Diff, which exhibits artifacts and loss of subject coherence as sparsity increases.
    The bottom row shows our Importance-Aware OBS, which better preserves the structural and semantic integrity of the generated image by prioritizing critical feature regions.}
    \label{fig:sd3_sample67}
    \vspace{-5mm}
\end{figure*}

\vspace{5mm}
\parag{SD3-Medium.} Table \ref{tab:1000_SD3_final} compares our approach to OBS-Diff on SD3-Medium and shows that injecting the CFG mask during pruning outperforms the baseline, particularly as sparsity levels increase. By leveraging semantic guidance from the CFG map, our method maintains higher alignment (CLIP Score) and human preference (ImageReward), preserving the model's generative quality even at 50\% sparsity, where standard OBS-Diff begins to exhibit structural degradation. As shown in Fig. \ref{fig:sd3_sample67}, our approach consistently preserves the object integrity and fine-grained semantic textures across various sparsity levels, even at an aggressive level of 60\%.

\begin{figure*}[!ht]
    \centering
    \small

    % Stable width control
    \setlength{\densew}{0.36\textwidth}      % dense column width
    \setlength{\denseimgw}{0.30\textwidth}   % actual dense image size
    \setlength{\imgw}{0.17\textwidth}        % each pruned image

    % ================= Dense column =================
    \begin{minipage}[c]{\densew}
        \centering
        {Dense}\\[0.8ex]
        \includegraphics[width=\denseimgw]{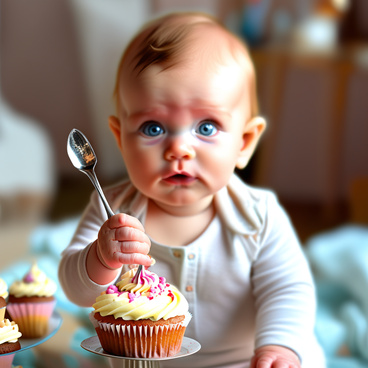}

        \vspace{1mm}
        {\footnotesize
        \textbf{Prompt:} ``A baby holding a spoon looking at a cupcake and candle.''}
    \end{minipage}
    \hspace{0.015\textwidth}
    % ================= Pruned grid =================
    \begin{minipage}[c]{0.60\textwidth}
        \centering

        % Row 1: OBS-Diff
        \includegraphics[width=\imgw]{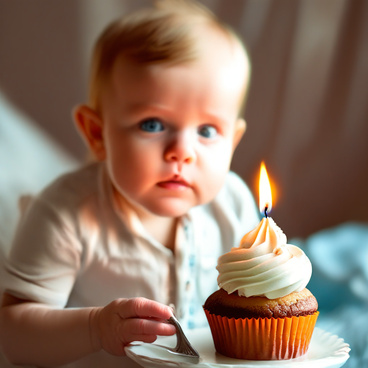}
        \hspace{3pt}
        \includegraphics[width=\imgw]{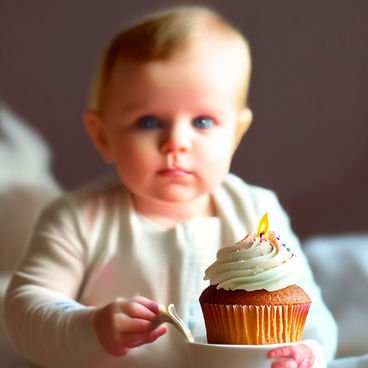}
        \hspace{3pt}
        \includegraphics[width=\imgw]{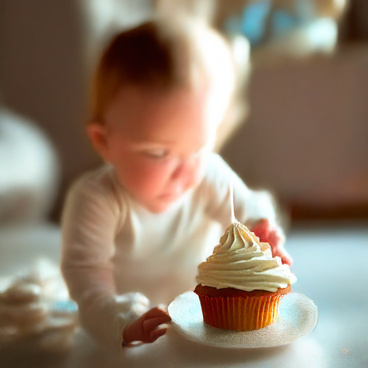}
        \hspace{2pt}
        \raisebox{0.45\imgw}{\rotatebox[origin=c]{270}{OBS-Diff~\cite{zhu2025obs}}}
        \\[2pt]

        % Row 2: Ours
        \includegraphics[width=\imgw]{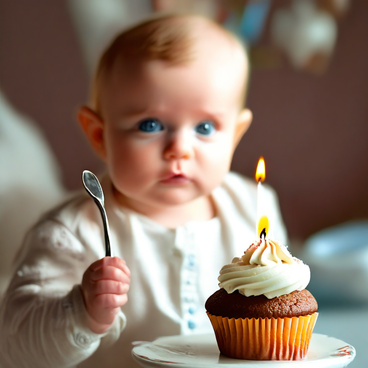}
        \hspace{3pt}
        \includegraphics[width=\imgw]{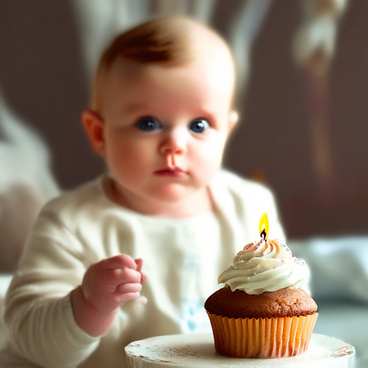}
        \hspace{3pt}
        \includegraphics[width=\imgw]{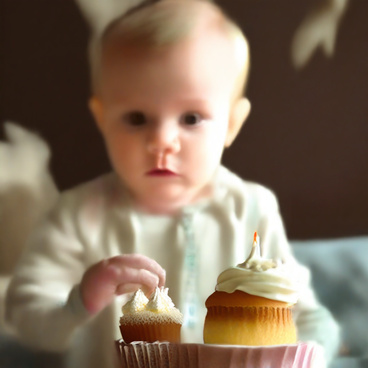}
        \hspace{2pt}
        \raisebox{0.45\imgw}{\rotatebox[origin=c]{270}{{Ours}}}
        % Column headers
        \makebox[\imgw][c]{50\% Sparsity}
        \hspace{3pt}
        \makebox[\imgw][c]{55\% Sparsity}
        \hspace{3pt}
        \makebox[\imgw][c]{60\% Sparsity}
        \hspace{0.025\textwidth}
        \\[0.8ex]

    \end{minipage}

    \vspace{2mm}
    \caption{\textbf{Qualitative comparison across different sparsity levels on PixArt-$\Sigma$ for Unstructured Pruning.}
    The top row shows OBS-Diff, which exhibits artifacts and loss of subject coherence as sparsity increases.
    The bottom row shows our Importance-Aware OBS, which better preserves the structural and semantic integrity of the generated image by prioritizing critical feature regions.}
    \label{fig:pix_513}
    \vspace{-5mm}
\end{figure*}

\parag{PixArt-$\Sigma$.}  As shown in Table \ref{tab:pix_cfg_vs_obs}, our approach to pruning with the CFG mask outperforms the baseline OBS-Diff on PixArt-$\Sigma$ across several aggressive sparsity levels. While both methods maintain high CLIP Scores, our method demonstrates superior resilience in human preference metrics (ImageReward). Notably, at 60\% sparsity, our approach achieves an ImageReward of $0.52$ compared to $0.49$ for OBS-Diff. Fig. \ref{fig:pix_513} further shows the accurately important semantic preservation of our approach relative to OBS-Diff. This indicates that by focusing the Hessian approximation on regions of high classifier-free guidance, we effectively prioritize the preservation of weights responsible for the primary semantic subjects of the prompt. %This allows the model to retain visual coherence and subject-level detail even when more than half of the parameters are removed, whereas the baseline tends to suffer from more generalized structural degradation. %suggesting that the inclusion of the CFG mask effectively preserves the semantic features that are the most important salient describing in the prompts (e.g., main objects), even when more than half of the model parameters are removed.

\begin{table}[!htp]
\centering
\caption{Quantitative comparison between OBS-Diff and Ours (CFG) on PixArt-$\Sigma$ using 1000 samples from GCC3M for unstructured pruning.}
\label{tab:pix_cfg_vs_obs}
\setlength{\tabcolsep}{0pt}
\begin{tabular*}{\textwidth}{@{\extracolsep{\fill}} lllccc}
\toprule
Model & Sparsity & Method & CLIP Score $\uparrow$ & ImageReward $\uparrow$ & MUSIQ $\uparrow$ \\ 
\midrule

\multirow{9}{*}{PixArt-$\Sigma$} 
& \multicolumn{2}{l}{Dense Model} & 31.86 & 0.94 & 71.11 \\ 
\cmidrule{2-6} 

& \multirow{2}{*}{45\%} 
& OBS-Diff & 31.63 & 0.78 & 70.12 \\
&  
& Ours (CFG) & \textbf{31.73} & \textbf{0.85} & \textbf{70.77} \\ 
\cmidrule{2-6} 

& \multirow{2}{*}{50\%} 
& OBS-Diff & 31.62 & 0.76 & 70.07 \\
&  
& Ours (CFG) & \textbf{31.71} & \textbf{0.78} & \textbf{70.34} \\ 
\cmidrule{2-6} 

& \multirow{2}{*}{55\%} 
& OBS-Diff & 31.65 & 0.64 & 68.25 \\
&  
& Ours (CFG) & \textbf{31.69} & \textbf{0.70} & \textbf{69.25} \\ 
\cmidrule{2-6} 

& \multirow{2}{*}{60\%} 
& OBS-Diff & 31.44 & 0.49 & 65.24 \\
&  
& Ours (CFG) & \textbf{31.54} & \textbf{0.52} & \textbf{67.13} \\ 
\bottomrule
\end{tabular*}
\end{table}

\subsection{Flexibility to Different Importance Guidance Signals} 

Our framework can use different importance signals without changing the pruning formulation. Besides the CFG mask, we evaluate Canny edge masks and object-local CFG masks. The Canny mask highlights image edges, while the object-local CFG mask is obtained by detecting objects with YOLOv8~\cite{10533619} and retaining the corresponding regions from the original CFG map.

% Tab.~\ref{tab:pix_canny_comparison} shows that different signals emphasize different preservation goals. CFG provides general semantic guidance, Canny preserves structural details, and object-local CFG focuses pruning on foreground objects. This distinction becomes clearer at high sparsity. At 60\% sparsity, Ours (CFG+ObjectDetector) achieves the best CLIP Score of $31.60$, while Ours (Canny) obtains the highest ImageReward of $0.53$, indicating stronger object-level alignment and better perceptual quality, respectively.

Tab.~\ref{tab:pix_canny_comparison} demonstrates the flexibility of our importance-aware pruning framework. The CFG signal can be replaced by a simple, model-agnostic cue such as Canny edges, making the approach readily applicable to diffusion models where CFG-based maps are unavailable or inconvenient to extract. With Canny guidance, the pruning objective shifts toward preserving finer structural details and local boundaries. Moreover, by incorporating an object detector on top of the CFG, our framework can be steered toward object-centric preservation. This customization yields consistent metric improvements and better maintains object structures at high sparsity.

Fig.~\ref{fig:flex_00000} confirms the same trend qualitatively. Object-local CFG better preserves the motorcycle parts (e.g., windshield and kickstand) by focusing on detected object regions, while Canny produces sharper wheel textures by emphasizing edge structure. These results show that our method is modular: different guidance signals can be plugged in to preserve different semantic or structural priors under pruning.

\begin{table}[!htp]
\centering
\caption{Quantitative comparison including Ours (Canny) and Ours (CFG+Detector) on PixArt-$\Sigma$ for Unstructured Pruning. Best results per sparsity level are bolded.}
\label{tab:pix_canny_comparison}
\setlength{\tabcolsep}{0pt}
\begin{tabular*}{\textwidth}{@{\extracolsep{\fill}} lllccc}
\toprule
\textbf{Model} & \textbf{Sparsity} & \textbf{Method} & \textbf{CLIP Score} $\uparrow$ & \textbf{ImageReward} $\uparrow$ & \textbf{MUSIQ} $\uparrow$ \\ \midrule
\multirow{19}{*}{PixArt-$\Sigma$} & \multicolumn{2}{l}{\textbf{Dense Model}} & 31.86 & 0.94 & 71.11 \\ \cmidrule{2-6} 
 & \multirow{4}{*}{45\%} & OBS-Diff & 31.63 & 0.78 & 70.12 \\
 &  & Ours (CFG) & \textbf{31.73} & \textbf{0.85} & 70.77 \\
 &  & {Ours (CFG+Detector)} & \textbf{31.73} & 0.82 & \textbf{70.87} \\
 &  & Ours (Canny) & \textbf{31.73} & 0.82 & 70.81 \\ \cmidrule{2-6} 
 & \multirow{4}{*}{50\%} & OBS-Diff & 31.62 & 0.76 & 70.07 \\
 &  & Ours (CFG) & 31.71 & 0.78 & 70.34 \\
 &  & {Ours (CFG+Detector)} & 31.71 & 0.78 & \textbf{70.59} \\
 &  & Ours (Canny) & \textbf{31.75} & \textbf{0.81} & 70.34 \\ \cmidrule{2-6} 
 & \multirow{4}{*}{55\%} & OBS-Diff & 31.65 & 0.64 & 68.25 \\
 &  & Ours (CFG) & 31.69 & 0.70 & 69.25 \\
 &  & {Ours (CFG+Detector)} & 31.79 & 0.75 & \textbf{69.62} \\
 &  & Ours (Canny) & \textbf{31.83} & \textbf{0.76} & 69.12 \\ \cmidrule{2-6} 
 & \multirow{4}{*}{60\%} & OBS-Diff & 31.44 & 0.49 & 65.24 \\
 &  & Ours (CFG) & 31.54 & 0.52 & 67.13 \\
 &  & {Ours (CFG+Detector)} & \textbf{31.60} & 0.50 & \textbf{67.27} \\
 &  & Ours (Canny) & 31.52 & \textbf{0.53} & 65.91 \\ \bottomrule
\end{tabular*}
\end{table}

\renewcommand{\arraystretch}{0.9}
\newcommand{\visPixArtFlexibility}[2]{
\begin{figure*}[!ht]
\centering
\small

\setlength{\tabcolsep}{2pt}

\def\imgw{0.203\textwidth}
\def\densew{0.315\textwidth}

\begin{minipage}{0.70\textwidth}
    \centering
    % Changed 'p{16pt}' to '>{\centering\arraybackslash}m{16pt}' for vertical/horizontal centering
    \begin{tabular}{@{} >{\centering\arraybackslash}m{16pt} cccc @{}}
        & OBS-Diff & Ours (CFG) & Ours (Det.) & Ours (Canny) \\[0.5ex]

        \rotatebox{90}{\textbf{Image}} &
        \adjustbox{valign=m}{\includegraphics[width=\imgw]{figures/Flexibility_PixArt/#1/images/OBS_Diff_55pct.jpg}} &
        \adjustbox{valign=m}{\includegraphics[width=\imgw]{figures/Flexibility_PixArt/#1/images/Ours_CFG_55pct.jpg}} &
        \adjustbox{valign=m}{\includegraphics[width=\imgw]{figures/Flexibility_PixArt/#1/images/Ours_CFG_Yolo_55pct_highlight.jpg}} &
        \adjustbox{valign=m}{\includegraphics[width=\imgw]{figures/Flexibility_PixArt/#1/images/Ours_Canny_55pct_highlight.jpg}} \\[6ex]

        \rotatebox{90}{\textbf{Mask}} &
        \adjustbox{valign=m}{\rule{0pt}{\imgw}} &
        \adjustbox{valign=m}{\includegraphics[width=\imgw]{figures/Flexibility_PixArt/#1/masks/CFG.jpg}} &
        \adjustbox{valign=m}{\includegraphics[width=\imgw]{figures/Flexibility_PixArt/#1/masks/Yolo.jpg}} &
        \adjustbox{valign=m}{\includegraphics[width=\imgw]{figures/Flexibility_PixArt/#1/masks/Canny.jpg}}
    \end{tabular}
\end{minipage}
% \hspace{0.5pt}
\hfill
\begin{minipage}{0.29\textwidth}
    \centering
    \textbf{Dense}\\[0.5ex]
    
    % \vspace{1.2em} % manual vertical alignment tweak
    \includegraphics[width=\linewidth]{figures/Flexibility_PixArt/#1/images/Original.jpg}
\end{minipage}

\vspace{2mm}
\caption{\textbf{Pruned Models with Different Important Signals.} Comparison of the OBS-Diff baseline against our importance-aware variants (CFG, CFG+Detector, and Canny). The second row visualizes the corresponding saliency masks. \textbf{Prompt:} "#2"}
\label{fig:flex_#1}
\vspace{-5mm}

\end{figure*}
}

\visPixArtFlexibility{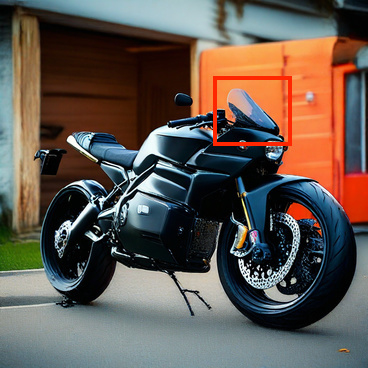}{black Honda motorcycle parked in front of a garage.}

% \visPixArtFlexibility{00620}{A large plant is in the corner of a small bathroom.}

%\subsection{Category-targeted Pruning} 
\subsection{Biasing Pruning Toward Targeted Categories}

% Our approach can effectively preserve the targeted predefined semantic subjects in pruning. To demonstrate this ability, we split the pruning and testing datasets into several category-targeted sub-datasets, where there are only prompts containing the specific category. Then, we adopt our approach and baseline to prune on these datasets. Besides comparing with the baseline, we also use our approach (general), which is pruned on the original datasets (containing category-diversity prompts), for comparison to highlight the effectiveness of our approach pruned with a predefined semantic category signal. Note that, these three competitors are tested on these category-targeted sub-datasets for comparison.

Our framework suggests a broader view of pruning: compression can be biased toward preserving user-specified properties rather than treating all visual content uniformly. We study category preservation as a simple instance of this idea. A straightforward way to introduce this bias is through the calibration data: We construct category-targeted pruning subsets whose prompts contain a predefined category, such as ``airplane'', ``cat'', and ``woman''. We compare OBS-Diff, our general model pruned on the original category-diverse calibration set, and our targeted model pruned on the corresponding category-specific subset. All models are evaluated on the same category-targeted test prompts.

\begin{figure*}[!htp]
    \centering
    \small
    \def\imgw{0.23\textwidth}
    \vspace{-2mm}
    \begin{tabular}{@{} c @{\hspace{5pt}} c @{\hspace{5pt}} c @{\hspace{5pt}} c @{}}
        % --- Headers ---
        {OBS-Diff (General)} & {OBS-Diff (Target)} & {Ours (General)} & {Ours (Target)} \\

        % --- Example: woman ---
        \adjustbox{valign=m}{\includegraphics[width=\imgw]{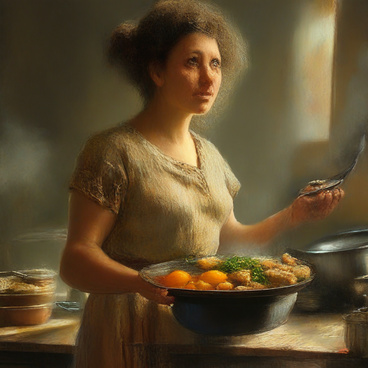}} & 
        \adjustbox{valign=m}{\includegraphics[width=\imgw]{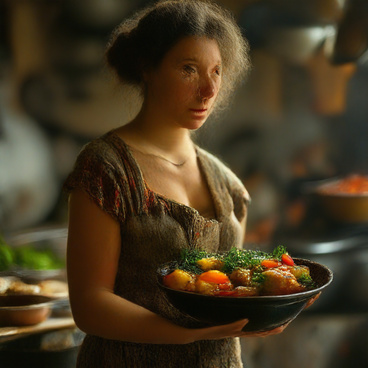}} & 
        \adjustbox{valign=m}{\includegraphics[width=\imgw]{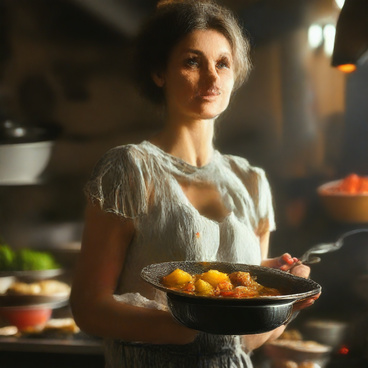}} & 
        \adjustbox{valign=m}{\includegraphics[width=\imgw]{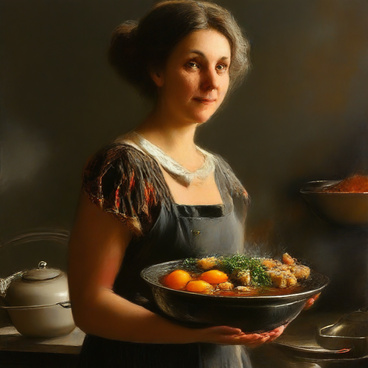}} \\
    \noalign{\vskip 4pt}
        % --- Example: cat ---
        \adjustbox{valign=m}{\includegraphics[width=\imgw]{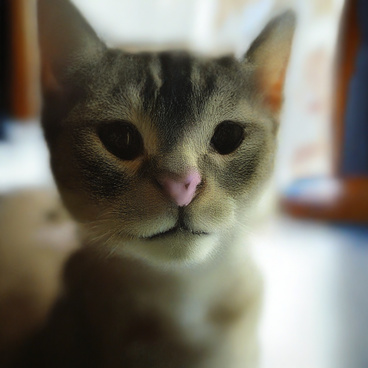}} & 
        \adjustbox{valign=m}{\includegraphics[width=\imgw]{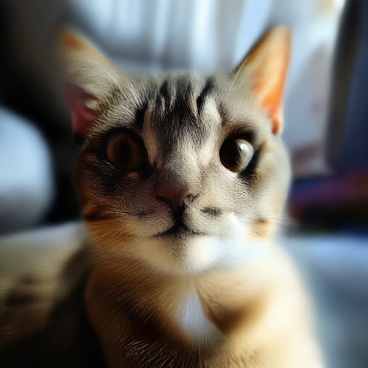}} & 
        \adjustbox{valign=m}{\includegraphics[width=\imgw]{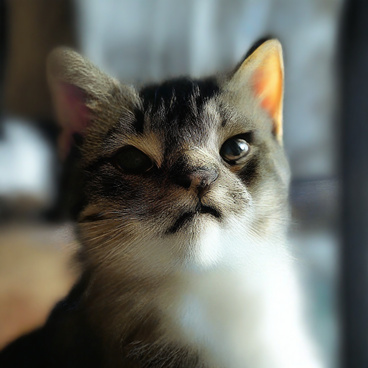}} & 
        \adjustbox{valign=m}{\includegraphics[width=\imgw]{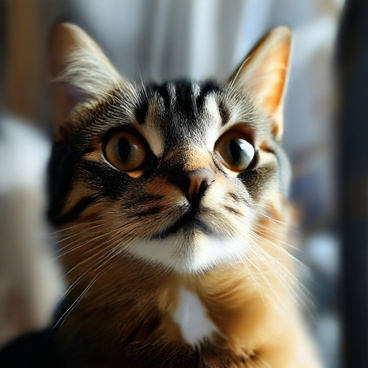}} \\
    \end{tabular}

    \caption{\textbf{Target-category pruning at 60\% sparsity.}
Category-targeted pruning improves preservation of the specified category, and our method further improves target fidelity and visual quality.}
    \label{fig:target_category_pruning}
    % \vspace{-5mm}
\end{figure*}

Tab.~\ref{tab:targeted_category_60pct_expanded} shows that category-specific calibration already steers pruning toward the target concept. At 60\% sparsity, Ours (target) outperforms OBS-Diff and our general variant on most quality metrics. For example, it improves MUSIQ from $69.52$ to $71.64$ on Cat and achieves the best MUSIQ score of $67.40$ on Woman. These results support our hypothesis that targeted calibration shifts the Hessian approximation and saliency estimates toward parameters important for the selected category. Additional spatial guidance further strengthens this bias, making targeted preservation more effective under high sparsity. Fig.\ref{fig:target_category_pruning} shows two visual examples for two categories - ``women'' and ``cat''. For more qualitative results, please refer to Appendix \ref{supp:cat_target_pruning}.

\begin{table}[h]
\centering
\vspace{-5mm}
\caption{Quantitative results of unstructured category-targeted pruning on PixArt-$\Sigma$ at 60\% sparsity.}
\label{tab:targeted_category_60pct_expanded}
\resizebox{0.8\textwidth}{!}{
\begin{tabular}{llccc}
\toprule
Category & Method & CLIP Score $\uparrow$ & ImageReward $\uparrow$ & MUSIQ $\uparrow$ \\ 
\midrule

\multirow{4}{*}{Woman}      
    & OBS-Diff (general) & 32.36 & 0.72 & 65.08 \\
    & OBS-Diff (target) & 32.25 & 0.75 & 65.75 \\
    & Ours (general)   & 32.31 & 0.76 & 66.74 \\
    & Ours (target)    & \textbf{32.38} & \textbf{0.85} & \textbf{67.40} \\ 
\midrule

\multirow{4}{*}{Cat}        
    & OBS-Diff (general) & 32.58 & 0.28 & 64.35 \\
    & OBS-Diff (target) & 32.45 & 0.29 & 69.81 \\
    & Ours (general)   & \textbf{32.66} & \textbf{0.41} & 69.52 \\
    & Ours (target)    & 32.28 & 0.34 & \textbf{71.64} \\ 
\midrule

% \multirow{4}{*}{Motorcycle} 
%     & OBS-Diff (general) & \textbf{31.94} & \textbf{1.07} & 66.50 \\
%     & OBS-Diff (target) & 30.88 & 0.92 & 68.07 \\
%     & Ours (general)   & 31.68 & 1.03 & 69.03 \\
%     & Ours (target)    & 30.92 & 0.99 & \textbf{71.66} \\ 
% \midrule

\multirow{4}{*}{Airplane}   
    & OBS-Diff (general) & \textbf{29.35} & 0.45 & 67.81 \\
    & OBS-Diff (target) & 28.62 & 0.43 & 69.90 \\
    & Ours (general)   & 29.22 & \textbf{0.62} & 68.19 \\
    & Ours (target)    & 28.76 & 0.51 & \textbf{70.81} \\ 
\bottomrule
\end{tabular}
}
\end{table}

\subsection{Structured Pruning}
Our formulation can be directly extended to structured pruning by following the aggregation strategy in OBS-Diff. Table~\ref{tab:pix_structured_cfg_vs_obs} shows that our method consistently outperforms OBS-Diff across sparsity levels in this setting. At 40\% sparsity, OBS-Diff drops to an ImageReward of $-0.04$, while our CFG-guided pruning achieves $0.30$, indicating better preservation of semantic and structural details. This is also reflected in \cref{fig:teaser} and \cref{fig:structured-airplane}, with additional qualitative results in Appendix~\ref{supp:structured_pruning}.

\begin{figure*}[!ht]
    \centering
    \small
    
    % Define image width (3 columns * 0.31 = ~0.93\textwidth, leaving space for labels)
    \def\imgw{0.28\textwidth}

    % 1 column for labels, 3 columns for images
    \begin{tabular}{@{} >{\centering\arraybackslash}m{15pt} c c c @{}}
        
        % --- Headers ---
        & \textbf{20\% Sparsity} & \textbf{30\% Sparsity} & \textbf{40\% Sparsity} \\[1ex]

        % --- Row 1: OBS-Diff ---
        \rotatebox{90}{\textbf{OBS-Diff}} &
        \adjustbox{valign=m}{\includegraphics[width=\imgw]{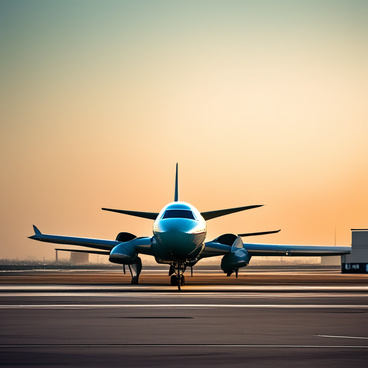}} &
        \adjustbox{valign=m}{\includegraphics[width=\imgw]{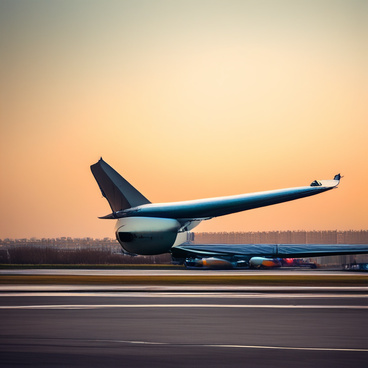}} &
        \adjustbox{valign=m}{\includegraphics[width=\imgw]{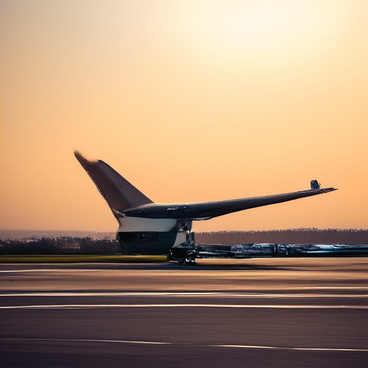}} \\[1.5ex] % Gap between rows

        % --- Row 2: Ours (CFG) ---
        \rotatebox{90}{\textbf{Ours}} &
        \adjustbox{valign=m}{\includegraphics[width=\imgw]{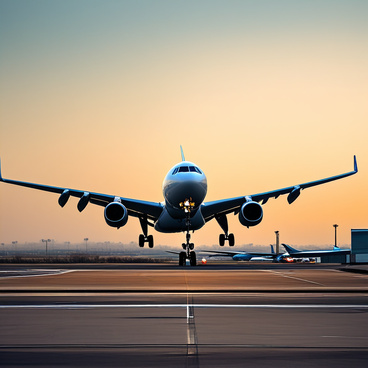}} &
        \adjustbox{valign=m}{\includegraphics[width=\imgw]{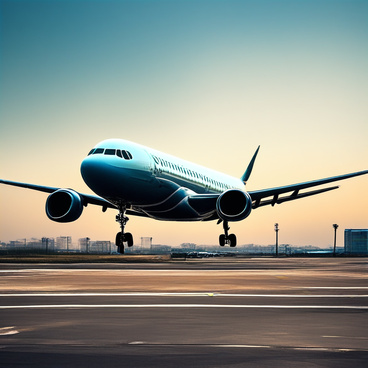}} &
        \adjustbox{valign=m}{\includegraphics[width=\imgw]{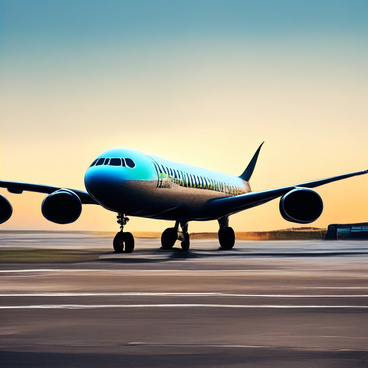}} \\
        
    \end{tabular}

    % \vspace{2mm}
    \caption{\textbf{Structured Pruning on PixArt-$\Sigma$.}
Comparison between OBS-Diff and our importance-aware pruning guided by CFG \cite{ho2022classifier} signals at 20\%, 30\%, and 40\% structured sparsity. \textbf{Prompt:} An Egyptian airlines plane landing at an airport.}
    \label{fig:structured-airplane}
\end{figure*}

\begin{table}[!ht]
\centering
\vspace{-5mm}
\caption{Quantitative comparison between OBS-Diff and Ours (CFG) on PixArt-$\Sigma$ using 1000 samples from GCC3M for structured pruning.}
\label{tab:pix_structured_cfg_vs_obs}
\setlength{\tabcolsep}{0pt}
\begin{tabular*}{\textwidth}{@{\extracolsep{\fill}} lllccc}
\toprule
Model & Sparsity & Method & CLIP Score $\uparrow$ & ImageReward $\uparrow$ & MUSIQ $\uparrow$ \\ 
\midrule

\multirow{9}{*}{PixArt-$\Sigma$} 
& \multicolumn{2}{l}{Dense Model} & 31.86 & 0.94 & 71.11 \\ 
\cmidrule{2-6} 

& \multirow{2}{*}{20\%} 
& OBS-Diff & 31.65 & 0.74 & 69.57 \\
&  
& Ours (CFG) & \textbf{31.68} & \textbf{0.79} & \textbf{71.37} \\ 
\cmidrule{2-6} 

& \multirow{2}{*}{30\%} 
& OBS-Diff & \textbf{31.36} & 0.38 & 66.98 \\
&  
& Ours (CFG) & 31.30 & \textbf{0.54} & \textbf{69.85} \\ 
\cmidrule{2-6} 

& \multirow{2}{*}{40\%} 
& OBS-Diff & 30.84 & -0.04 & 65.17 \\
&  
& Ours (CFG) & \textbf{31.08} & \textbf{0.30} & \textbf{68.64} \\ 
\bottomrule
\end{tabular*}
\end{table}

%------------------------------------------------

\subsection{Ablation Study}
\label{ssec:ablation}

We conduct an ablation study on PixArt-$\Sigma$ to analyze the impact of the Classifier-Free Guidance (CFG) scale during the pruning phase.

\parag{The impact of the guidance scale.}

We analyze the sensitivity of mask extraction to the Classifier-Free Guidance (CFG) scale used during the pruning phase. Specifically, we compare pruning PixArt-$\Sigma$ under guidance scales of $\text{GS} = 4.5$ and $\text{GS} = 7.0$, while keeping the evaluation guidance scale fixed at $7.0$. As shown in Table~\ref{tab:merged_pruning_gs_blue}, our method consistently outperforms OBS-Diff across both $45\%$ and $50\%$ sparsity levels under both guidance configurations.

While OBS-Diff exhibits performance degradation when reducing the pruning CFG scale from $7.0$ to $4.5$ (e.g., CLIP Score drops from $31.71$ to $31.63$ at $45\%$ sparsity), our approach maintains robust text alignment (CLIP Score $\ge 31.71$) across all settings. Furthermore, at $\text{GS} = 4.5$, our method achieves its highest ImageReward of $0.85$ at $45\%$ sparsity and $0.78$ at $50\%$ sparsity. This demonstrates that our importance-guided saliency estimation effectively preserves critical parameter weights without relying on high guidance amplification during mask extraction.

\begin{table}[h]
\centering
\caption{Quantitative results of unstructured pruning of PixArt-$\Sigma$ at 45\% and 50\% sparsity levels, comparing pruning guidance scales of 4.5 and 7.0 (tested at 7.0). The best results per guidance scale are highlighted in bold.}
\label{tab:merged_pruning_gs_blue}
\setlength{\tabcolsep}{0pt}
\begin{tabular*}{\textwidth}{@{\extracolsep{\fill}} llcccc}
\toprule
& & \multicolumn{2}{c}{\textbf{Guidance Scale = 4.5}} & \multicolumn{2}{c}{\textbf{Guidance Scale = 7.0}} \\ \cmidrule(lr){3-4} \cmidrule(lr){5-6}
\textbf{Sparsity} & \textbf{Method} & \textbf{CLIP Score} $\uparrow$ & \textbf{ImageReward} $\uparrow$ & \textbf{CLIP Score} $\uparrow$ & \textbf{ImageReward} $\uparrow$ \\ \midrule
\multicolumn{2}{l}{\textbf{Dense Model}} & 31.86 & 0.94 & 31.86 & 0.94 \\ \cmidrule{1-6}
\multirow{2}{*}{45\%} & OBS-Diff & 31.63 & 0.78 & 31.71 & 0.79 \\
                      & {Ours} & \textbf{31.73} & \textbf{0.85} & \textbf{31.72} & \textbf{0.82} \\ \cmidrule{1-6}
\multirow{2}{*}{50\%} & OBS-Diff & 31.62 & 0.76 & 31.70 & 0.76 \\
                      & {Ours} & \textbf{31.71} & \textbf{0.78} & \textbf{31.73} & \textbf{0.77} \\ \bottomrule
\end{tabular*}
\end{table}

\subsection{Human Evaluation}
To further assess perceptual quality, we conduct a user study in which participants rate generated images on a five-point scale from 1 (collapsed) to 5 (excellent). As shown in Tab.~\ref{tab:human_comparison}, our method is strongly preferred over OBS-Diff, achieving a substantially higher human score of 3.97 compared to 2.59 at 60\% sparsity. Notably, despite this clear improvement in human-perceived quality, our method receives a slightly lower CLIPScore. This discrepancy further highlights that automatic metrics do not always reliably reflect perceptual image quality. We provide additional analysis and qualitative examples in Appendix \ref{appendix:user-study}.

\begin{table}[!htp]
\centering
\caption{Quantitative Comparison on 20-Sample Evaluation Set on PixArt-$\Sigma$ at 60\% Sparsity. Bold values indicate superior performance in automated metrics.}
\label{tab:human_comparison}
\begin{tabular}{lcccc}
\toprule
Method & CLIP Score $\uparrow$ & ImageReward $\uparrow$ & MUSIQ $\uparrow$ & Human Score $\uparrow$ \\ 
\midrule
OBS-Diff & \textbf{31.90} & 0.72 & 69.83 & 2.59 \\
Ours     & 31.06 & \textbf{0.78} & \textbf{72.93} & \textbf{3.97} \\ 
\bottomrule
\end{tabular}
\end{table}

\section{Limitations}
Our framework inherits the limitations of the chosen importance signal. CFG-based masks are training-free, but they can be noisy or incomplete for abstract prompts, small objects, and heavily cluttered scenes. In addition, our method is designed as a post-training pruning approach and does not explore whether finetuning can recover further quality at very high sparsity. For unstructured pruning, practical latency gains still require hardware or sparse-kernel support. Finally, our evaluation is centered on text-to-image generation; extending the same principle to video diffusion, editing, personalization, and other conditional generation tasks remains future work.
\section{Conclusion}
We introduced an importance-aware pruning framework for diffusion models that shifts compression from uniformly preserving all reconstruction errors to preserving the errors that matter most for generation. By incorporating semantic importance signals into second-order pruning, our method better protects prompt-relevant regions and fine-grained structures without requiring retraining or architectural changes. Across unstructured, structured, and category-targeted pruning settings, our approach consistently improves semantic preservation over OBS-Diff, especially under aggressive sparsity where baseline pruning often collapses subject structure. These results show that diffusion pruning should not be treated as a purely parameter-level compression problem: effective compression must account for where and what the model needs to preserve. We believe this work supports a broader view of guided model compression, where pruning objectives can be biased toward user-specified semantic, spatial, or task-level priorities.

\textbf{{Acknowledgments}} This work was supported by faculty startup funds from the College of Computing and Informatics (CCI) at the University of North Carolina at Charlotte.

\FloatBarrier
\bibliographystyle{IEEEtran}
\bibliography{main, haoyu}

%%%%%%%%%%%%%%%%%%%%%%%%%%%%%%%%%%%%%%%%%%%%%%%%%%%%%%%%%%%%
\newpage
\appendix

% \section{Discussions.}

\section{Evaluation metrics and User Study.}
\label{appendix:user-study}
Evaluating compressed text-to-image models remains challenging because automatic metrics only partially capture the failures introduced by pruning. CLIP-based metrics are useful for measuring coarse image-text alignment, but they are often insensitive to localized semantic damage, object deformation, missing fine details, and spatial relation errors. This limitation is consistent with recent text-to-image evaluation studies showing that CLIPScore often behaves as a global alignment measure and can fail on compositional prompts involving objects, attributes, and relations~\cite{lin2024evaluating,ghosh2023geneval}. This weakness is especially relevant in our setting: pruning does not always destroy the entire image, but often damages foreground subjects, object boundaries, limbs, faces, or other semantically important regions. Such failures can still receive reasonable CLIP scores as long as the image remains broadly related to the prompt.

We therefore include ImageReward and MUSIQ as complementary metrics. ImageReward provides a stronger preference-oriented signal because it is trained from human preference annotations and is designed to better reflect perceptual quality and prompt fidelity in text-to-image generation~\cite{xu2023imagereward}. MUSIQ, on the other hand, measures image quality without relying directly on text-image alignment, making it useful for detecting visual degradation such as blur, artifacts, and unnatural structure. In our benchmark results, ImageReward and MUSIQ generally separate visually coherent generations from degraded ones more clearly than CLIPScore, especially at high sparsity. However, both remain automatic metrics and should not be treated as complete substitutes for human judgment.

For this reason, we further conduct a user study to directly measure perceived image quality. Each participant rates every generated image on a five-level quality scale: 1-\emph{Collapsed} for incoherent, broken, or unusable images; 2-\emph{Distorted} for images with major defects and poor visual quality; 3-\emph{Degraded} for images with visible artifacts but still recognizable content; 4-\emph{Good} for mostly natural images with minor artifacts; and 5-\emph{Excellent} for clean, natural, artifact-free images. This protocol allows us to compare automatic metric preferences against human judgments at the individual-sample level.

\begin{figure*}[!htp]
    \centering
    
    % --- First Misaligned Example (ID 00093, Index 17) ---
    \begin{subfigure}{\textwidth}
        \centering
        % Images side-by-side
        \begin{minipage}{0.48\textwidth}
            \centering
            \includegraphics[width=\linewidth]{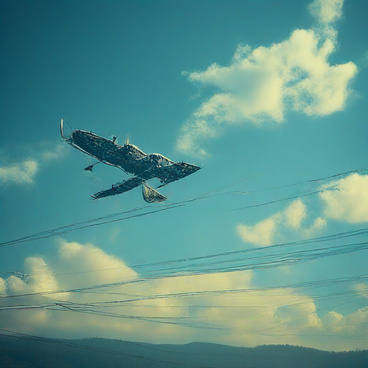} 
            \caption*{OBS-Diff}
        \end{minipage}
        \hfill
        \begin{minipage}{0.48\textwidth}
            \centering
            \includegraphics[width=\linewidth]{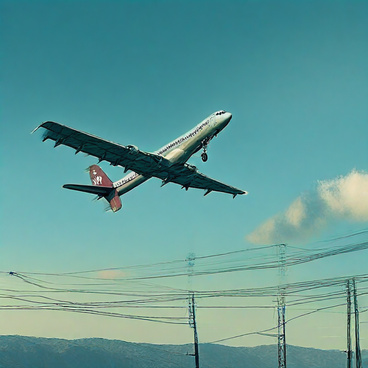} 
            \caption*{Ours (CFG)}
        \end{minipage}
        
        \vspace{2mm}
        
        % Mini-table for scores
        \resizebox{0.75\textwidth}{!}{
        \begin{tabular}{lcccc}
        \toprule
        Method & Human Score $\uparrow$ & CLIP Score $\uparrow$ & ImageReward $\uparrow$ & MUSIQ $\uparrow$ \\
        \midrule
        OBS-Diff & 1.65 & \textbf{29.71} & 1.34 & 73.58 \\
        Ours     & \textbf{3.59} & 28.79 & \textbf{1.60} & \textbf{74.46} \\
        \bottomrule
        \end{tabular}
        }
        \vspace{1mm}
        \caption{Sample A (Index: 17). Human evaluators strongly prefer our method ($3.59$ vs $1.65$). ImageReward and MUSIQ correctly align with the human preference by scoring our method higher, whereas the CLIP Score incorrectly favors the baseline. \textbf{Prompt:} "An airplane high in the sky over some electrical wires."}
        \label{fig:misalign_a}
        \vspace{6mm}
    \end{subfigure}
    
    % --- Second Misaligned Example (ID 00173, Index 7) ---
    \begin{subfigure}{\textwidth}
        \centering
        % Images side-by-side
        \begin{minipage}{0.48\textwidth}
            \centering
            \includegraphics[width=\linewidth]{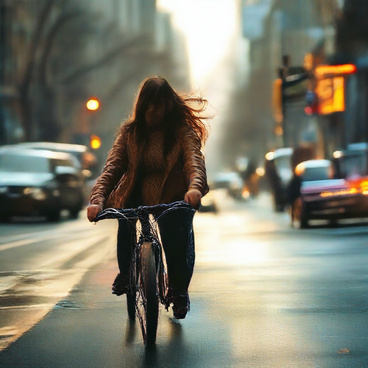} 
            \caption*{OBS-Diff}
        \end{minipage}
        \hfill
        \begin{minipage}{0.48\textwidth}
            \centering
            \includegraphics[width=\linewidth]{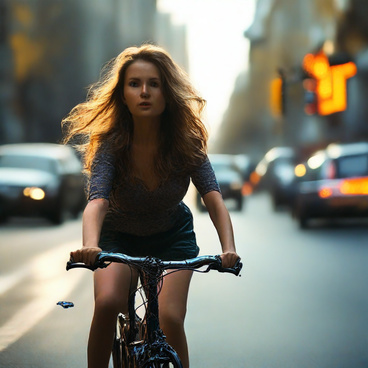} 
            \caption*{Ours (CFG)}
        \end{minipage}
        
        \vspace{2mm}
        
        % Mini-table for scores
        \resizebox{0.75\textwidth}{!}{
        \begin{tabular}{lcccc}
        \toprule
        Method & Human Score $\uparrow$ & CLIP Score $\uparrow$ & ImageReward $\uparrow$ & MUSIQ $\uparrow$ \\
        \midrule
        OBS-Diff & 2.29 & \textbf{29.30} & 1.71 & 65.07 \\
        Ours     & \textbf{3.59} & 28.34 & \textbf{1.93} & \textbf{70.41} \\
        \bottomrule
        \end{tabular}
        }
        \vspace{1mm}
        \caption{Sample B (Index: 7). Human evaluators again show a strong preference for our method ($3.59$ vs $2.29$). While ImageReward and MUSIQ successfully align with human perception, the CLIP Score incorrectly assigns a higher value to the baseline. \textbf{Prompt:} "A woman is riding her bike down the street in front of traffic".}
        \label{fig:misalign_b}
    \end{subfigure}
    
    \caption{Visual comparisons highlighting the misalignment between automated metrics and human perception at 60\% sparsity. Bold values indicate the higher score. These examples demonstrate the limitations of CLIP Score in reliably capturing human-perceived visual quality following pruning, a gap bridged more effectively by ImageReward and MUSIQ.}
    \label{fig:metric_misalignment}
\end{figure*}

Our results in Tab.\ref{tab:detailed_human_comparison} show that ImageReward and MUSIQ align better with human ratings than CLIPScore, but none of the automatic metrics is perfect. Qualitative examples reveal cases where CLIPScore remains similar, or even favors the baseline, despite clear degradation in subject integrity, edge structure, or object-level detail, as can be seen in Fig.\ref{fig:metric_misalignment}. These observations support our main evaluation choice: the benefit of importance-aware pruning should not be judged solely by global embedding-based metrics. Its advantage is most visible in preserving the semantically important regions that humans notice first, but that CLIP-style metrics may under-penalize.

% \begin{table}[!htp]
% \centering
% \caption{Quantitative Comparison on 20-Sample Evaluation Set on PixArt-$\Sigma$ at 60\% Sparsity. Bold values indicate superior performance in automated metrics.}
% \label{tab:human_comparison}
% \begin{tabular}{lcccc}
% \toprule
% Method & CLIP Score $\uparrow$ & ImageReward $\uparrow$ & MUSIQ $\uparrow$ & Human Score $\uparrow$ \\ 
% \midrule
% OBS-Diff & \textbf{31.90} & 0.72 & 69.83 & 2.59 \\
% Ours     & 31.06 & \textbf{0.78} & \textbf{72.93} & \textbf{3.97} \\ 
% \bottomrule
% \end{tabular}
% \end{table}

\begin{table}[!htp]
\centering
\caption{Detailed per-sample alignment between mean human scores and automated metric preferences. \cmark indicates the metric preference matches the human preference ($H_O > H_B$), while \xmark indicates the metric preference contradicts the human preference.}
\label{tab:detailed_human_comparison}
\small
\begin{tabular}{ccc|ccc}
\toprule
Index & Human (B) & Human (O) & CLIP Score & ImageReward & MUSIQ \\
\midrule
1  & 2.00 & \textbf{3.59} & \xmark & \xmark & \cmark \\
2  & 2.53 & \textbf{4.06} & \cmark & \xmark & \cmark \\
3  & 2.59 & \textbf{3.29} & \xmark & \xmark & \cmark \\
4  & 2.41 & \textbf{3.18} & \xmark & \xmark & \cmark \\
\midrule
5  & 2.24 & \textbf{3.76} & \cmark & \cmark & \xmark \\
6  & 2.18 & \textbf{4.00} & \cmark & \xmark & \cmark \\
7  & 2.29 & \textbf{3.59} & \xmark & \cmark & \cmark \\
8  & 2.53 & \textbf{3.59} & \xmark & \cmark & \cmark \\
\midrule
9  & 3.00 & \textbf{4.59} & \xmark & \xmark & \cmark \\
10 & 3.35 & \textbf{4.18} & \xmark & \xmark & \xmark \\
11 & 2.88 & \textbf{4.59} & \xmark & \xmark & \xmark \\
12 & 3.00 & \textbf{4.53} & \xmark & \cmark & \cmark \\
\midrule
13 & 3.18 & \textbf{4.56} & \xmark & \cmark & \cmark \\
14 & 3.47 & \textbf{4.65} & \cmark & \cmark & \xmark \\
15 & 2.65 & \textbf{4.76} & \xmark & \xmark & \cmark \\
16 & 2.94 & \textbf{4.24} & \xmark & \cmark & \cmark \\
\midrule
17 & 1.65 & \textbf{3.59} & \xmark & \cmark & \cmark \\
18 & 2.53 & \textbf{3.76} & \xmark & \cmark & \xmark \\
19 & 2.12 & \textbf{3.59} & \xmark & \cmark & \cmark \\
20 & 2.18 & \textbf{3.29} & \cmark & \xmark & \xmark \\
\midrule
\multicolumn{3}{l|}{\textbf{Average Alignment With Human Score}} & 25.00\% & 50.00\% & 70.00\% \\
\bottomrule
\end{tabular}
\end{table}

\FloatBarrier

\section{Qualitative Results}

\subsection{Results on PixArt-$\Sigma$}
Fig. \ref{fig:pixart_00037} - Fig. \ref{fig:pixart_00036} illustrate qualitative comparison across aggressively increasing sparsity using PixArt-$\Sigma$ with various prompts. We compare our approach using the original CFG mask as an important guidance signal to the OBS-Diff baseline.

\subsection{Results on SD3-Medium}
Fig. \ref{fig:SD3_00389} - Fig. \ref{fig:SD3_00377} illustrate qualitative comparison across aggressively increasing sparsity using SD3-Medium with various prompts. We compare our approach using the original CFG mask as an important guidance signal to the OBS-Diff baseline.

\subsection{Flexibility Across Different Importance Signals} 
\label{supp:flex_pruning}
Fig. \ref{fig:flexibility_00000} - Fig. \ref{fig:flexibility_00610} illustrate comparison of the baseline OBS-Diff against our importance-aware variants guided by different signals (CFG, Object Detector, Canny Edge) at 45\%, 50\%, 55\%, and 60\% unstructured sparsity.

\subsection{Biasing Pruning Toward Targeted Categories} 
\label{supp:cat_target_pruning}

%\input{tabs/targetcat}
%Table %\ref{tab:targeted_category_60pct_expanded} \HL{TODO}

% \paragraph{Chair}

\begin{figure*}[!htp]
    \centering
    \small
    \def\imgw{0.23\textwidth}

    \begin{tabular}{@{} c @{\hspace{5pt}} c @{\hspace{5pt}} c @{\hspace{5pt}} c @{}}
        % --- Headers ---
        \textbf{OBS-Diff (General)} & \textbf{OBS-Diff (Target)} & \textbf{Ours (General)} & \textbf{Ours (Target)} \\[1ex]
        % --- Example: 00181 ---
        \adjustbox{valign=m}{\includegraphics[width=\imgw]{figures/Target_pruning/woman/00181/OBS-Diff_General_60pct.jpg}} & 
        \adjustbox{valign=m}{\includegraphics[width=\imgw]{figures/Target_pruning/woman/00181/OBS-Diff_60pct.jpg}} & 
        \adjustbox{valign=m}{\includegraphics[width=\imgw]{figures/Target_pruning/woman/00181/Ours_General_60pct.jpg}} & 
        \adjustbox{valign=m}{\includegraphics[width=\imgw]{figures/Target_pruning/woman/00181/Ours_Target_60pct.jpg}} \\[1ex]
        % --- Example: 00195 ---
        \adjustbox{valign=m}{\includegraphics[width=\imgw]{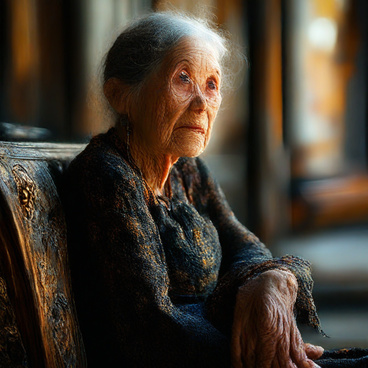}} & 
        \adjustbox{valign=m}{\includegraphics[width=\imgw]{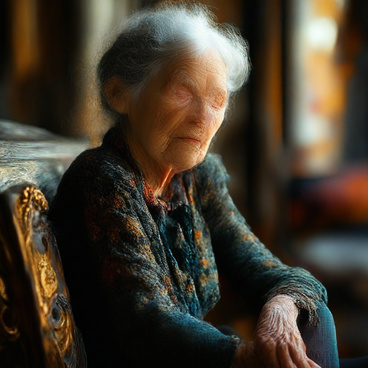}} & 
        \adjustbox{valign=m}{\includegraphics[width=\imgw]{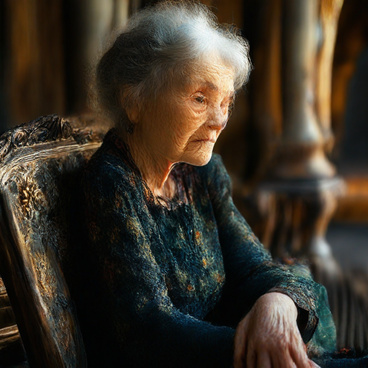}} & 
        \adjustbox{valign=m}{\includegraphics[width=\imgw]{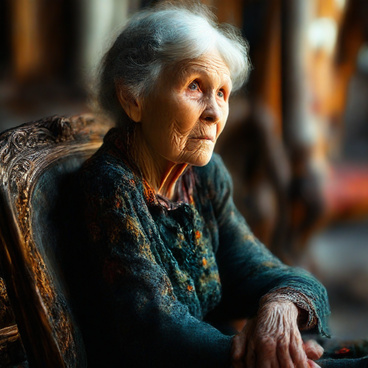}} \\[1ex]
        % --- Example: 00173 ---
        \adjustbox{valign=m}{\includegraphics[width=\imgw]{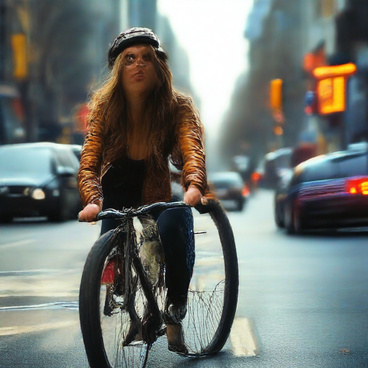}} & 
        \adjustbox{valign=m}{\includegraphics[width=\imgw]{figures/Target_pruning/woman/00173/OBS-Diff_60pct.jpg}} & 
        \adjustbox{valign=m}{\includegraphics[width=\imgw]{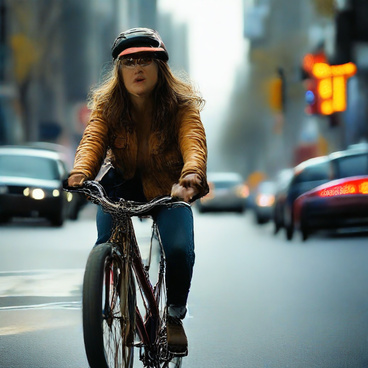}} & 
        \adjustbox{valign=m}{\includegraphics[width=\imgw]{figures/Target_pruning/woman/00173/Ours_Target_60pct.jpg}} \\[1ex]
        % --- Example: 00323 ---
        \adjustbox{valign=m}{\includegraphics[width=\imgw]{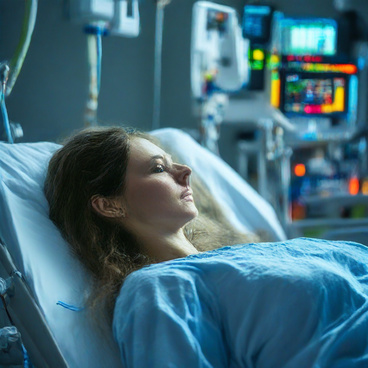}} & 
        \adjustbox{valign=m}{\includegraphics[width=\imgw]{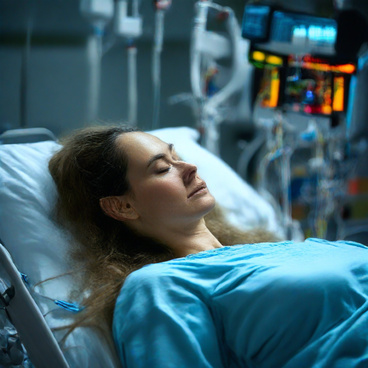}} & 
        \adjustbox{valign=m}{\includegraphics[width=\imgw]{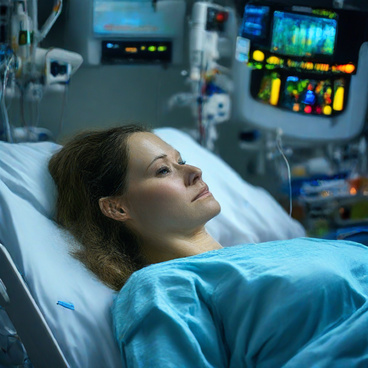}} & 
        \adjustbox{valign=m}{\includegraphics[width=\imgw]{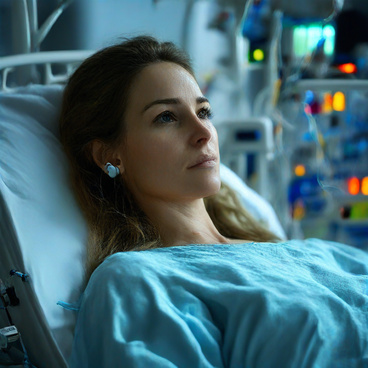}} \\
    \end{tabular}

    \vspace{2mm}
    \caption{\textbf{Target-Category Pruning: woman.} Comparison of general and targeted pruning baselines (OBS-Diff) against our importance-aware variants at 60\% sparsity level.}
    \label{fig:target_woman}
\end{figure*}

\paragraph{Woman}
Fig. \ref{fig:target_woman} shows a qualitative comparison among Ours (Target), Ours (General), and OBS-Diff (Target) at $60\%$ sparsity level for woman-targeted pruning.

\begin{figure*}[!ht]
    \centering
    \small
    \def\imgw{0.23\textwidth}

    \begin{tabular}{@{} c @{\hspace{5pt}} c @{\hspace{5pt}} c @{\hspace{5pt}} c @{}}
        % --- Headers ---
        \textbf{OBS-Diff (General)} & \textbf{OBS-Diff (Target)} & \textbf{Ours (General)} & \textbf{Ours (Target)} \\[1ex]
        % --- Example: 00032 ---
        \adjustbox{valign=m}{\includegraphics[width=\imgw]{figures/Target_pruning/cat/00032/OBS-Diff_General_60pct.jpg}} & 
        \adjustbox{valign=m}{\includegraphics[width=\imgw]{figures/Target_pruning/cat/00032/OBS-Diff_60pct.jpg}} & 
        \adjustbox{valign=m}{\includegraphics[width=\imgw]{figures/Target_pruning/cat/00032/Ours_General_60pct.jpg}} & 
        \adjustbox{valign=m}{\includegraphics[width=\imgw]{figures/Target_pruning/cat/00032/Ours_Target_60pct.jpg}} \\[1ex]
        % --- Example: 00065 ---
        \adjustbox{valign=m}{\includegraphics[width=\imgw]{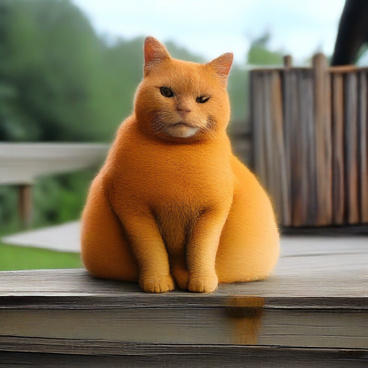}} & 
        \adjustbox{valign=m}{\includegraphics[width=\imgw]{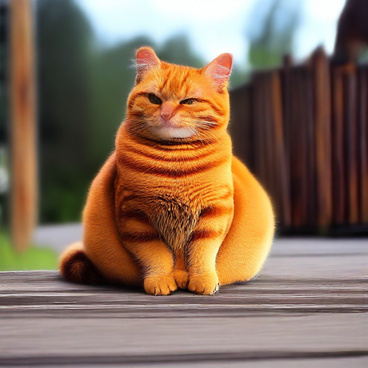}} & 
        \adjustbox{valign=m}{\includegraphics[width=\imgw]{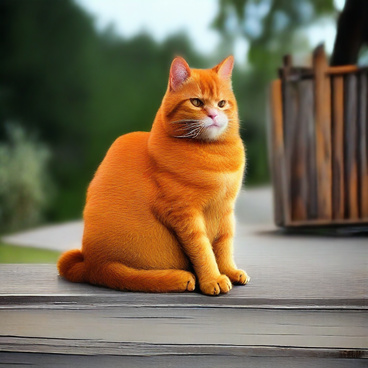}} & 
        \adjustbox{valign=m}{\includegraphics[width=\imgw]{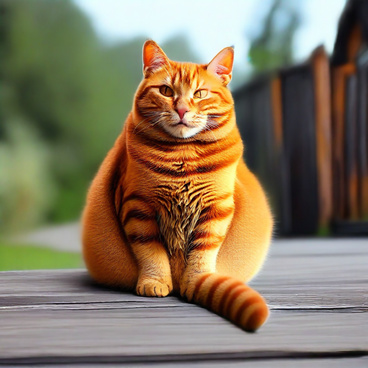}} \\[1ex]
        % --- Example: 00019 ---
        \adjustbox{valign=m}{\includegraphics[width=\imgw]{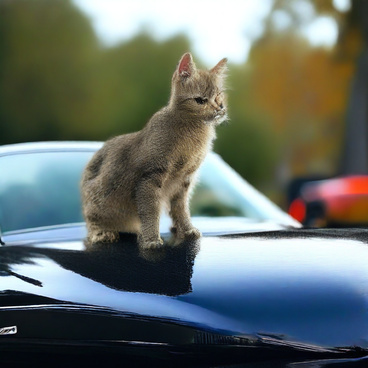}} & 
        \adjustbox{valign=m}{\includegraphics[width=\imgw]{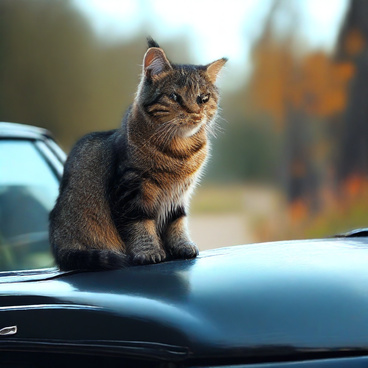}} & 
        \adjustbox{valign=m}{\includegraphics[width=\imgw]{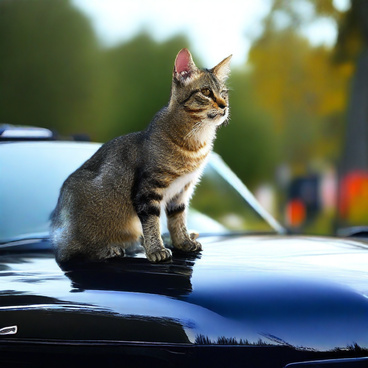}} & 
        \adjustbox{valign=m}{\includegraphics[width=\imgw]{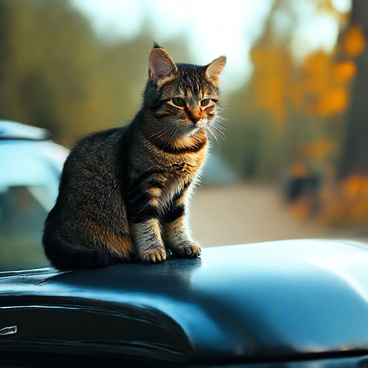}} \\[1ex]
        % --- Example: 00069 ---
        \adjustbox{valign=m}{\includegraphics[width=\imgw]{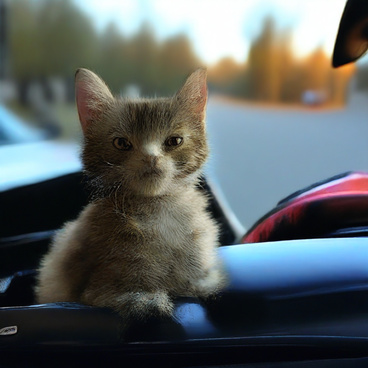}} & 
        \adjustbox{valign=m}{\includegraphics[width=\imgw]{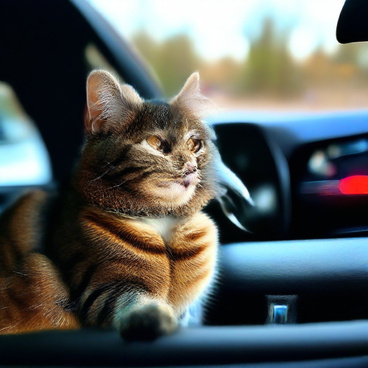}} & 
        \adjustbox{valign=m}{\includegraphics[width=\imgw]{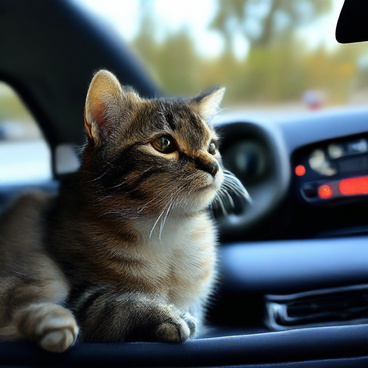}} & 
        \adjustbox{valign=m}{\includegraphics[width=\imgw]{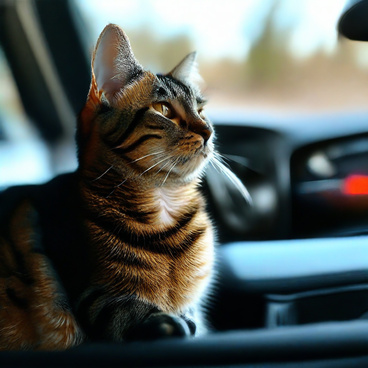}} \\
    \end{tabular}

    \vspace{2mm}
    \caption{\textbf{Target-Category Pruning: cat.} Comparison of general and targeted pruning baselines (OBS-Diff) against our importance-aware variants at 60\% sparsity level.}
    \label{fig:target_cat}
\end{figure*}

\paragraph{Cat}
Fig. \ref{fig:target_cat} shows a qualitative comparison among Ours (Target), Ours (General), and OBS-Diff (Target) at $60\%$ sparsity level for cat-targeted pruning.

\begin{figure*}[!ht]
    \centering
    \small
    \def\imgw{0.23\textwidth}

    \begin{tabular}{@{} c @{\hspace{5pt}} c @{\hspace{5pt}} c @{\hspace{5pt}} c @{}}
        % --- Headers ---
        \textbf{OBS-Diff (General)} & \textbf{OBS-Diff (Target)} & \textbf{Ours (General)} & \textbf{Ours (Target)} \\[1ex]
        % --- Example: 00009 ---
        \adjustbox{valign=m}{\includegraphics[width=\imgw]{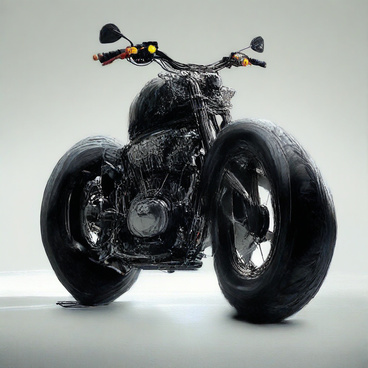}} & 
        \adjustbox{valign=m}{\includegraphics[width=\imgw]{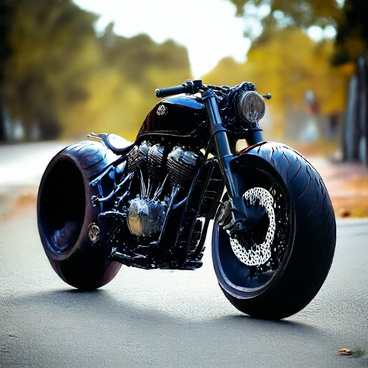}} & 
        \adjustbox{valign=m}{\includegraphics[width=\imgw]{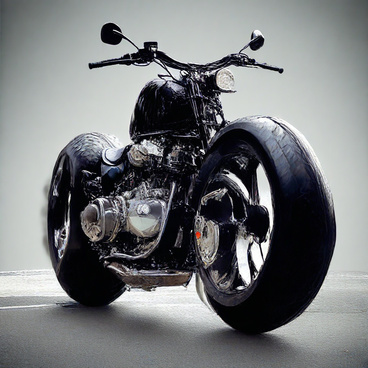}} & 
        \adjustbox{valign=m}{\includegraphics[width=\imgw]{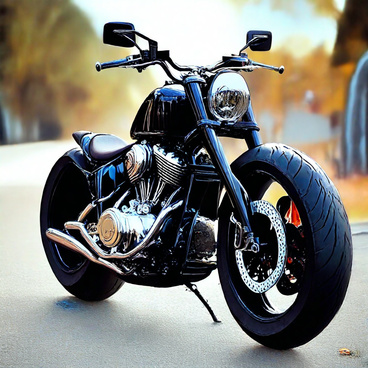}} \\[1ex]
        % --- Example: 00047 ---
        \adjustbox{valign=m}{\includegraphics[width=\imgw]{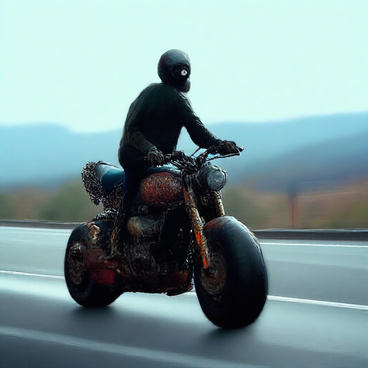}} & 
        \adjustbox{valign=m}{\includegraphics[width=\imgw]{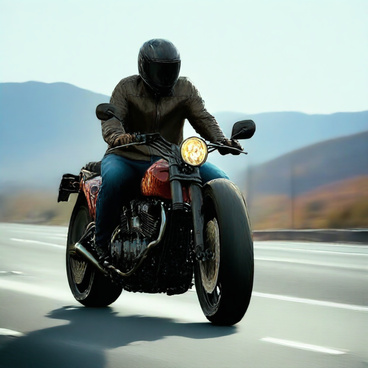}} & 
        \adjustbox{valign=m}{\includegraphics[width=\imgw]{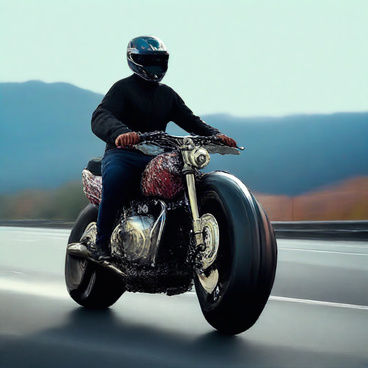}} & 
        \adjustbox{valign=m}{\includegraphics[width=\imgw]{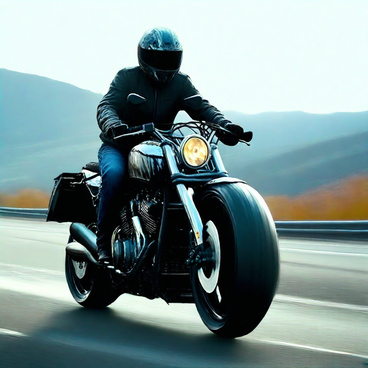}} \\[1ex]
        % --- Example: 00167 ---
        \adjustbox{valign=m}{\includegraphics[width=\imgw]{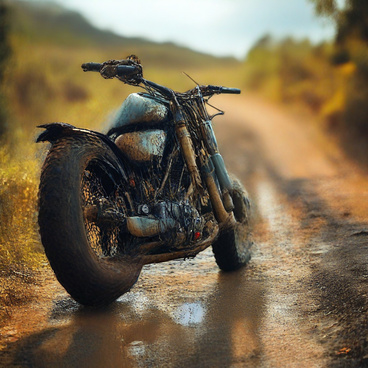}} & 
        \adjustbox{valign=m}{\includegraphics[width=\imgw]{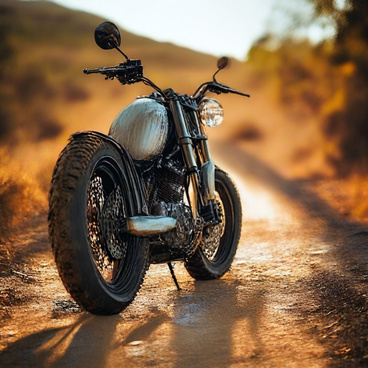}} & 
        \adjustbox{valign=m}{\includegraphics[width=\imgw]{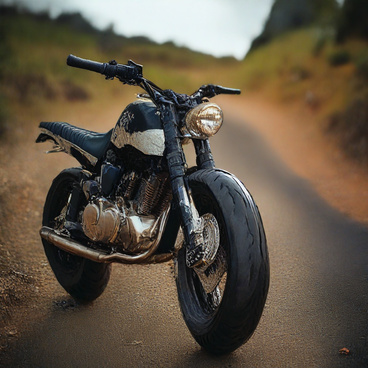}} & 
        \adjustbox{valign=m}{\includegraphics[width=\imgw]{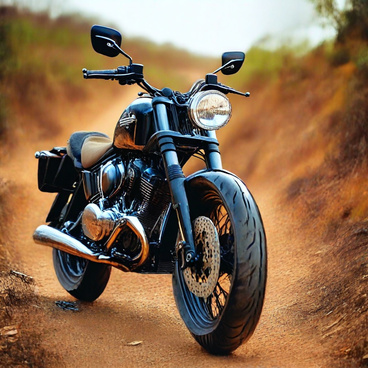}} \\[1ex]
        % --- Example: 00137 ---
        \adjustbox{valign=m}{\includegraphics[width=\imgw]{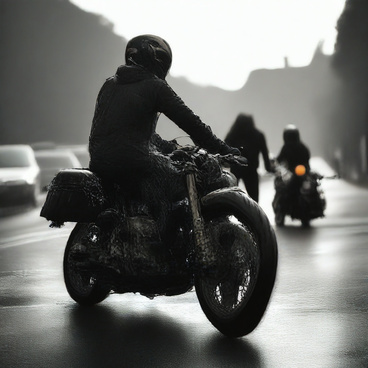}} & 
        \adjustbox{valign=m}{\includegraphics[width=\imgw]{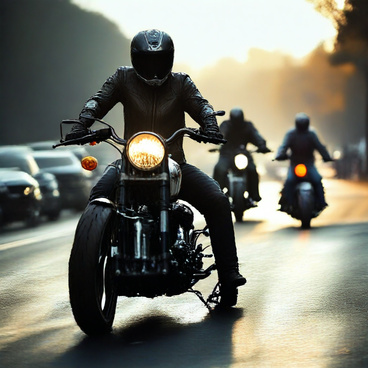}} & 
        \adjustbox{valign=m}{\includegraphics[width=\imgw]{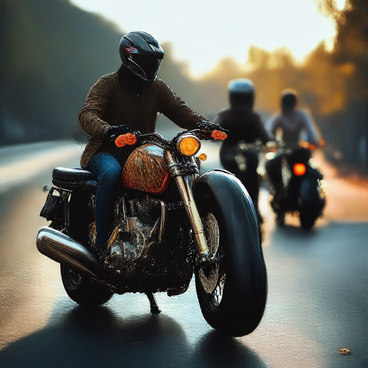}} & 
        \adjustbox{valign=m}{\includegraphics[width=\imgw]{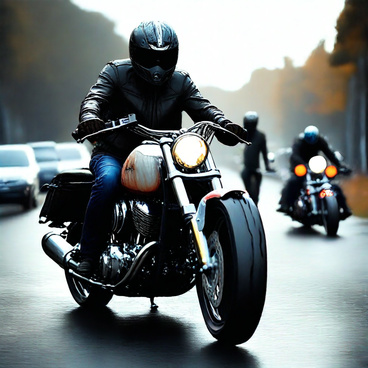}} \\
    \end{tabular}

    \vspace{2mm}
    \caption{\textbf{Target-Category Pruning: motorcycle.} Comparison of general and targeted pruning baselines (OBS-Diff) against our importance-aware variants at 60\% sparsity level.}
    \label{fig:target_motorcycle}
\end{figure*}

\paragraph{Motorcycle}
Fig. \ref{fig:target_motorcycle} shows a qualitative comparison among Ours (Target), Ours (General), and OBS-Diff (Target) at $60\%$ sparsity level for motorcycle-targeted pruning.

\begin{figure*}[!ht]
    \centering
    \small
    \def\imgw{0.23\textwidth}

    \begin{tabular}{@{} c @{\hspace{5pt}} c @{\hspace{5pt}} c @{\hspace{5pt}} c @{}}
        % --- Headers ---
        \textbf{OBS-Diff (General)} & \textbf{OBS-Diff (Target)} & \textbf{Ours (General)} & \textbf{Ours (Target)} \\[1ex]
        % --- Example: 00093 ---
        \adjustbox{valign=m}{\includegraphics[width=\imgw]{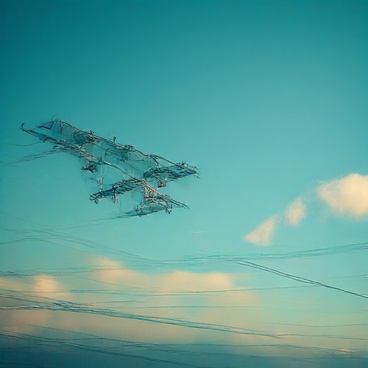}} & 
        \adjustbox{valign=m}{\includegraphics[width=\imgw]{figures/Target_pruning/airplane/00093/OBS-Diff_60pct.jpg}} & 
        \adjustbox{valign=m}{\includegraphics[width=\imgw]{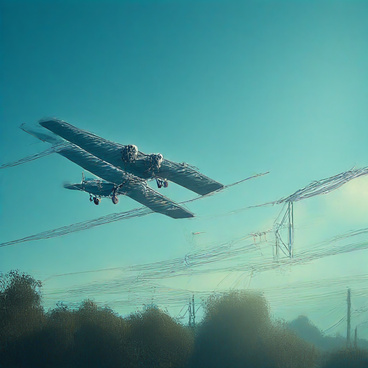}} & 
        \adjustbox{valign=m}{\includegraphics[width=\imgw]{figures/Target_pruning/airplane/00093/Ours_Target_60pct.jpg}} \\[1ex]
        % --- Example: 00031 ---
        \adjustbox{valign=m}{\includegraphics[width=\imgw]{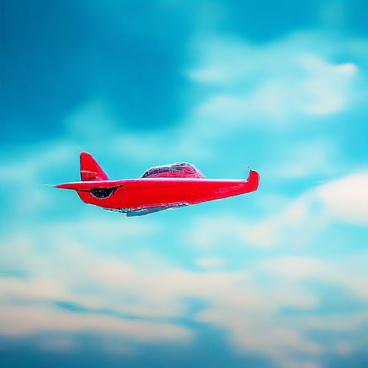}} & 
        \adjustbox{valign=m}{\includegraphics[width=\imgw]{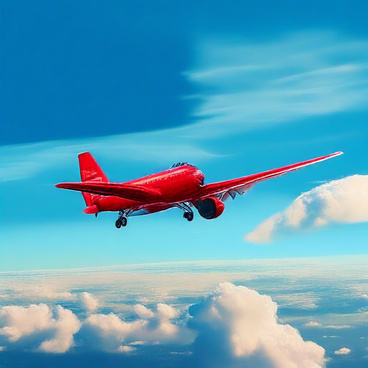}} & 
        \adjustbox{valign=m}{\includegraphics[width=\imgw]{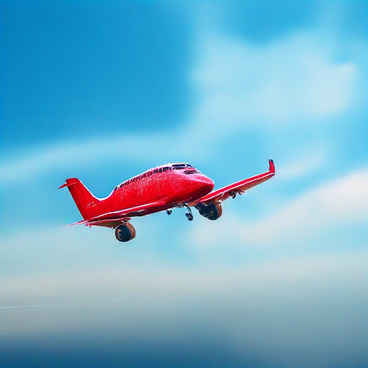}} & 
        \adjustbox{valign=m}{\includegraphics[width=\imgw]{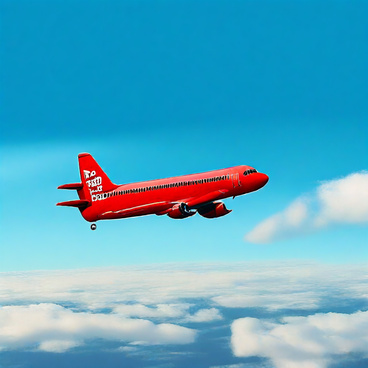}} \\[1ex]
        % --- Example: 00035 ---
        \adjustbox{valign=m}{\includegraphics[width=\imgw]{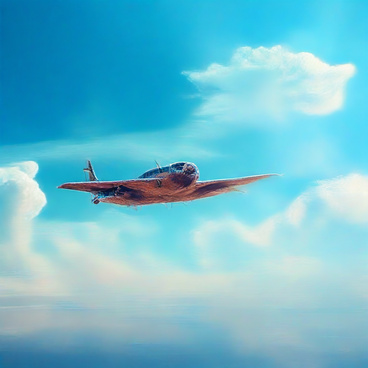}} & 
        \adjustbox{valign=m}{\includegraphics[width=\imgw]{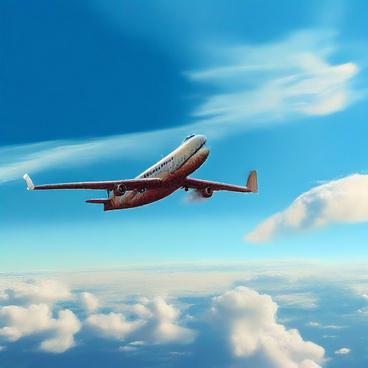}} & 
        \adjustbox{valign=m}{\includegraphics[width=\imgw]{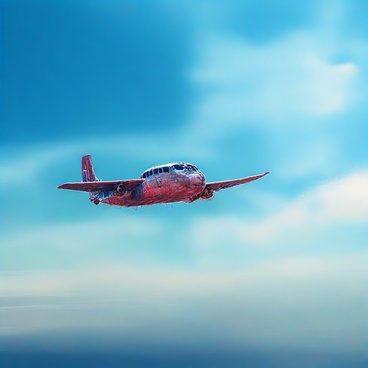}} & 
        \adjustbox{valign=m}{\includegraphics[width=\imgw]{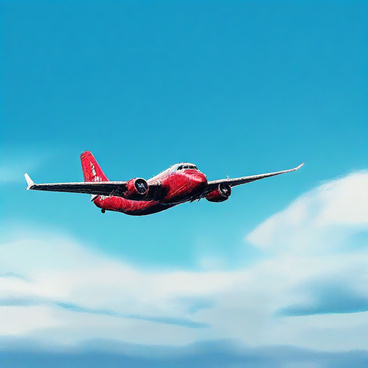}} \\[1ex]
        % --- Example: 00039 ---
        \adjustbox{valign=m}{\includegraphics[width=\imgw]{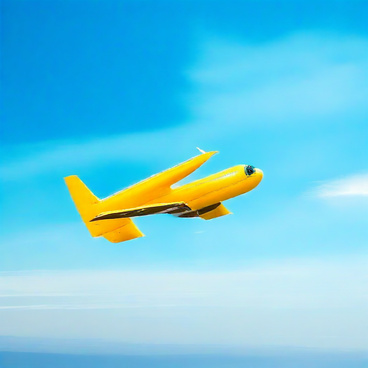}} & 
        \adjustbox{valign=m}{\includegraphics[width=\imgw]{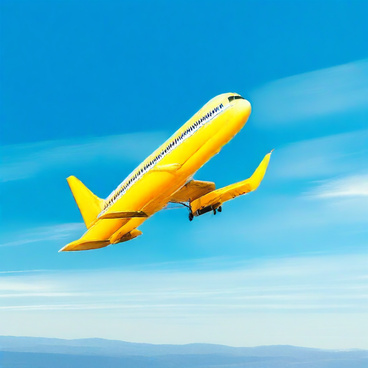}} & 
        \adjustbox{valign=m}{\includegraphics[width=\imgw]{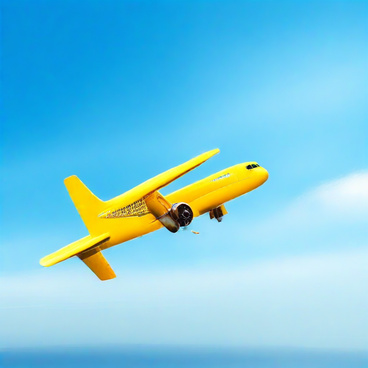}} & 
        \adjustbox{valign=m}{\includegraphics[width=\imgw]{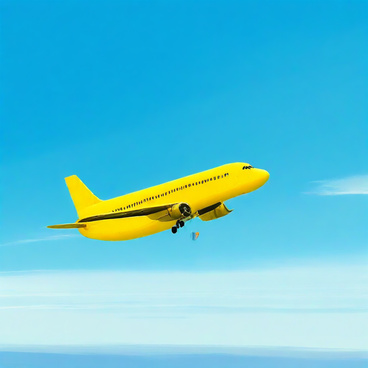}} \\
    \end{tabular}

    \vspace{2mm}
    \caption{\textbf{Target-Category Pruning: airplane.} Comparison of general and targeted pruning baselines (OBS-Diff) against our importance-aware variants at 60\% sparsity level.}
    \label{fig:target_airplane}
\end{figure*}

\paragraph{Airplane}
Fig. \ref{fig:target_airplane} shows a qualitative comparison among Ours (Target), Ours (General), and OBS-Diff (Target) at $60\%$ sparsity level for airplane-targeted pruning.

\subsection{Structured Pruning} 
\label{supp:structured_pruning}

Fig. \ref{fig:structured_00000} - Fig. \ref{fig:structured_00127} illustrate qualitative comparison across aggressively increasing sparsity using PixArt-$\Sigma$ with various prompts for Structured Pruning. We compare our approach using the original CFG mask as an important guidance signal to the OBS-Diff baseline.

\FloatBarrier
\section{Derivation of Importance-Aware OBS Saliency}
\label{app:derivation}

We derive the closed-form saliency score used by our importance-aware OBS
formulation. For layer $l$, let $X'_{l,t}=M_t\odot X_{l,t}$ denote the
importance-filtered activation at timestep $t$. The layer-wise reconstruction
objective is
\begin{equation}
\mathcal{J}(\hat W_l)
=
\sum_{t=1}^{T} \alpha_t
\left\|
\hat W_l X'_{l,t} - W_l X'_{l,t}
\right\|_F^2 .
\end{equation}
This objective is minimized at $\hat W_l=W_l$, where the reconstruction error
and the first-order gradient are both zero. Therefore, the second-order
expansion around $W_l$ is exact for this quadratic surrogate objective.

Following standard OBS-style layer-wise pruning, we derive the update for one
output row at a time. Let $w\in\mathbb{R}^{C_{\mathrm{in}}}$ denote one row of
$W_l$, and let $\delta w$ denote its perturbation after pruning. The
importance-aware Hessian for this row is
\begin{equation}
H_{l,\mathrm{imp}}
=
2\sum_{t=1}^{T}\alpha_t
\mathbb{E}\left[
X'_{l,t} X_{l,t}^{\prime\top}
\right],
\end{equation}
where the expectation is taken over the calibration data. The change in the
layer-wise reconstruction objective is then
\begin{equation}
\Delta \mathcal{J}
=
\frac{1}{2}
\delta w^\top H_{l,\mathrm{imp}} \delta w .
\end{equation}

To compute the saliency of weight $w_q$, we set this weight to zero while
allowing the remaining weights in the row to compensate. The pruning constraint
is
\begin{equation}
\mathbf{e}_q^\top \delta w + w_q = 0,
\end{equation}
where $\mathbf{e}_q$ is the unit vector selecting coordinate $q$. We minimize
$\Delta \mathcal{J}$ subject to this constraint using the Lagrangian
\begin{equation}
\mathcal{G}(\delta w,\lambda)
=
\frac{1}{2}
\delta w^\top H_{l,\mathrm{imp}} \delta w
+
\lambda
\left(
\mathbf{e}_q^\top \delta w + w_q
\right).
\end{equation}
Setting the first-order conditions to zero gives
\begin{align}
\frac{\partial \mathcal{G}}{\partial \delta w}=0
&\implies
H_{l,\mathrm{imp}}\delta w + \lambda \mathbf{e}_q = 0,
\\
\frac{\partial \mathcal{G}}{\partial \lambda}=0
&\implies
\mathbf{e}_q^\top \delta w + w_q = 0.
\end{align}
Thus,
\begin{equation}
\delta w
=
-\lambda H_{l,\mathrm{imp}}^{-1}\mathbf{e}_q .
\end{equation}
Substituting this into the constraint yields
\begin{equation}
-\lambda
\mathbf{e}_q^\top
H_{l,\mathrm{imp}}^{-1}
\mathbf{e}_q
+
w_q
=
0,
\end{equation}
and therefore
\begin{equation}
\lambda
=
\frac{w_q}
{
\left[
H_{l,\mathrm{imp}}^{-1}
\right]_{qq}
}.
\end{equation}
The optimal compensating update is
\begin{equation}
\delta w
=
-
\frac{w_q}
{
\left[
H_{l,\mathrm{imp}}^{-1}
\right]_{qq}
}
H_{l,\mathrm{imp}}^{-1}\mathbf{e}_q .
\end{equation}
The resulting saliency score is
\begin{equation}
S_{l,q}
=
\Delta \mathcal{J}
=
\frac{1}{2}
\delta w^\top H_{l,\mathrm{imp}}\delta w
=
\frac{w_q^2}
{
2
\left[
H_{l,\mathrm{imp}}^{-1}
\right]_{qq}
}.
\end{equation}

This derivation has the same structure as classical OBS
\citep{hassibi1992second}. The difference is that the Hessian is computed from
importance-filtered activations $X'_{l,t}$ rather than raw activations
$X_{l,t}$. Thus, the saliency estimate remains an OBS-style reconstruction
criterion, but it is biased toward preserving weights that affect spatially
important regions.
\newcommand{\visPixArt}[2]{
\begin{figure*}[!htp]
    \centering
    \small
    % --- Column Headers ---
    \begin{minipage}{0.03\linewidth} \vfill \end{minipage} 
    \begin{minipage}{0.23\linewidth} \centering 45\% Sparsity \end{minipage} \hfill
    \begin{minipage}{0.23\linewidth} \centering 50\% Sparsity \end{minipage} \hfill
    \begin{minipage}{0.23\linewidth} \centering 55\% Sparsity \end{minipage} \hfill
    \begin{minipage}{0.23\linewidth} \centering 60\% Sparsity \end{minipage}
    
    \vspace{1mm}

    % --- Row 1: Classical OBS-DIFF ---
    \begin{minipage}{0.03\linewidth}
        \rotatebox{90}{\centering \textbf{OBS-Diff}}
    \end{minipage}
    \begin{minipage}{0.96\linewidth}
        \begin{subfigure}[b]{0.24\linewidth}
            \centering \includegraphics[width=\linewidth]{figures/PixArt/#1/OBS-Diff_45pct.jpg}
        \end{subfigure} \hfill
        \begin{subfigure}[b]{0.24\linewidth}
            \centering \includegraphics[width=\linewidth]{figures/PixArt/#1/OBS-Diff_50pct.jpg}
        \end{subfigure} \hfill
        \begin{subfigure}[b]{0.24\linewidth}
            \centering \includegraphics[width=\linewidth]{figures/PixArt/#1/OBS-Diff_55pct.jpg}
        \end{subfigure} \hfill
        \begin{subfigure}[b]{0.24\linewidth}
            \centering \includegraphics[width=\linewidth]{figures/PixArt/#1/OBS-Diff_60pct.jpg}
        \end{subfigure}
    \end{minipage}

    \vspace{1mm}

    % --- Row 2: Ours ---
    \begin{minipage}{0.03\linewidth}
        \rotatebox{90}{\centering \textbf{Ours}}
    \end{minipage}
    \begin{minipage}{0.96\linewidth}
        \begin{subfigure}[b]{0.24\linewidth}
            \centering \includegraphics[width=\linewidth]{figures/PixArt/#1/Ours_CFG_45pct.jpg}
        \end{subfigure} \hfill
        \begin{subfigure}[b]{0.24\linewidth}
            \centering \includegraphics[width=\linewidth]{figures/PixArt/#1/Ours_CFG_50pct.jpg}
        \end{subfigure} \hfill
        \begin{subfigure}[b]{0.24\linewidth}
            \centering \includegraphics[width=\linewidth]{figures/PixArt/#1/Ours_CFG_55pct.jpg}
        \end{subfigure} \hfill
        \begin{subfigure}[b]{0.24\linewidth}
            \centering \includegraphics[width=\linewidth]{figures/PixArt/#1/Ours_CFG_60pct.jpg}
        \end{subfigure}
    \end{minipage}

    \caption{\textbf{Qualitative results on PixArt-$\Sigma$.}
Comparison between OBS-Diff and our importance-aware pruning guided by CFG \cite{ho2022classifier} signals at 45\%, 50\%, 55\%, and 60\% unstructured sparsity. \textbf{Prompt:} "#2"}
    \label{fig:pixart_#1}
\end{figure*}
% \clearpage % Optional: ensures figures don't pile up on one page
}
% 00037, 00156, 00000, 00261, 00100, 00248, 00008, 00012, 00035, 00036
\visPixArt{00037}{A young man and his cute cat enjoy a nap together.}
% A kitten sitting in a sink with a green brush with green bristles
\visPixArt{00513}{A kitten sitting in a sink with a green brush with green bristles.}

\visPixArt{00000}{A black Honda motorcycle parked in front of a garage.}

\visPixArt{00261}{A motorcycle is parked on a dirt road in a forest.}

\visPixArt{00100}{A customized motorcycle with a large rear and skinny front tire.}

\visPixArt{00248}{a shop a car traffic lights and buildings.}

% \visPixArt{00001}{A Honda motorcycle parked in a grass driveway.}

\visPixArt{00008}{A beautiful dessert waiting to be shared by two people.}

\visPixArt{00012}{A cat eating a bird it has caught.}

\visPixArt{00035}{A cat at attention between two parked cars.}

\visPixArt{00036}{A dog sitting between its masters feet on a footstool watching tv.}

\newcommand{\visSD}[2]{
\begin{figure*}[!htp]
    \centering
    \small
    % --- Column Headers ---
    \begin{minipage}{0.03\linewidth} \vfill \end{minipage} 
    \begin{minipage}{0.23\linewidth} \centering 30\% Sparsity \end{minipage} \hfill
    \begin{minipage}{0.23\linewidth} \centering 40\% Sparsity \end{minipage} \hfill
    \begin{minipage}{0.23\linewidth} \centering 45\% Sparsity \end{minipage} \hfill
    \begin{minipage}{0.23\linewidth} \centering 50\% Sparsity \end{minipage}
    
    \vspace{1mm}

    % --- Row 1: Classical OBS-DIFF ---
    \begin{minipage}{0.03\linewidth}
        \rotatebox{90}{\centering \textbf{OBS-Diff}}
    \end{minipage}
    \begin{minipage}{0.96\linewidth}
        \begin{subfigure}[b]{0.24\linewidth}
            \centering \includegraphics[width=\linewidth]{figures/SD3/#1/OBS-Diff_30pct.jpg}
        \end{subfigure} \hfill
        \begin{subfigure}[b]{0.24\linewidth}
            \centering \includegraphics[width=\linewidth]{figures/SD3/#1/OBS-Diff_40pct.jpg}
        \end{subfigure} \hfill
        \begin{subfigure}[b]{0.24\linewidth}
            \centering \includegraphics[width=\linewidth]{figures/SD3/#1/OBS-Diff_45pct.jpg}
        \end{subfigure} \hfill
        \begin{subfigure}[b]{0.24\linewidth}
            \centering \includegraphics[width=\linewidth]{figures/SD3/#1/OBS-Diff_50pct.jpg}
        \end{subfigure}
    \end{minipage}

    \vspace{1mm}

    % --- Row 2: Ours ---
    \begin{minipage}{0.03\linewidth}
        \rotatebox{90}{\centering \textbf{Ours}}
    \end{minipage}
    \begin{minipage}{0.96\linewidth}
        \begin{subfigure}[b]{0.24\linewidth}
            \centering \includegraphics[width=\linewidth]{figures/SD3/#1/Ours_30pct.jpg}
        \end{subfigure} \hfill
        \begin{subfigure}[b]{0.24\linewidth}
            \centering \includegraphics[width=\linewidth]{figures/SD3/#1/Ours_40pct.jpg}
        \end{subfigure} \hfill
        \begin{subfigure}[b]{0.24\linewidth}
            \centering \includegraphics[width=\linewidth]{figures/SD3/#1/Ours_45pct.jpg}
        \end{subfigure} \hfill
        \begin{subfigure}[b]{0.24\linewidth}
            \centering \includegraphics[width=\linewidth]{figures/SD3/#1/Ours_50pct.jpg}
        \end{subfigure}
    \end{minipage}

    \caption{\textbf{Qualitative resutls on SD3-Medium.}
Comparison between OBS-Diff and our importance-aware pruning guided by CFG \cite{ho2022classifier} signals at 30\%, 40\%, 45\%, and 50\% unstructured sparsity. \textbf{Prompt:} "#2"}
    \label{fig:SD3_#1}
\end{figure*}
}

\visSD{00389}{A man walking beside sheep on a country road.}

\visSD{00067}{An American Airlines plane is in the sky.}

% \visSD{00321}{A young man and his cute cat enjoy a nap together.}

\visSD{00323}{This new fridge goes great in this clean kitchen.}

\visSD{00326}{Two people riding a motorcycle to the beach.}

\visSD{00377}{A silver car in the street next to a metal railing.}


% Structured Pruning

\visStructured{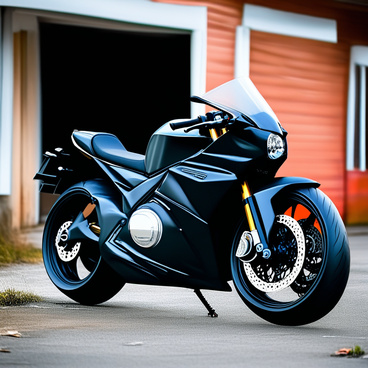}{A black Honda motorcycle parked in front of a garage.}
\visStructured{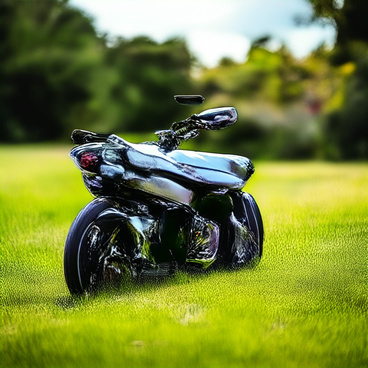}{A Honda motorcycle parked in a grass driveway.}
\visStructured{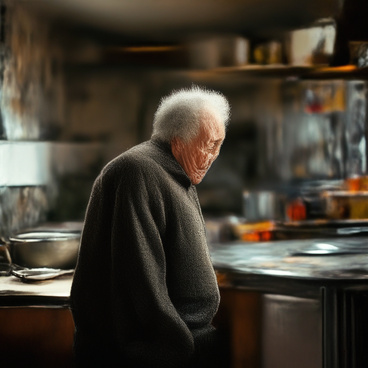}{A shot of an elderly man inside a kitchen.}
\visStructured{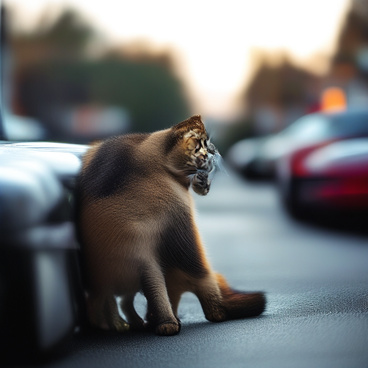}{A cat in between two cars in a parking lot.}
\visStructured{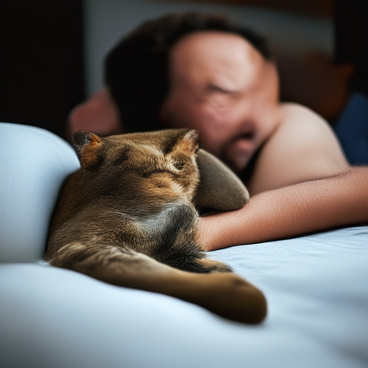}{a man sleeping with his cat next to him}
\visStructured{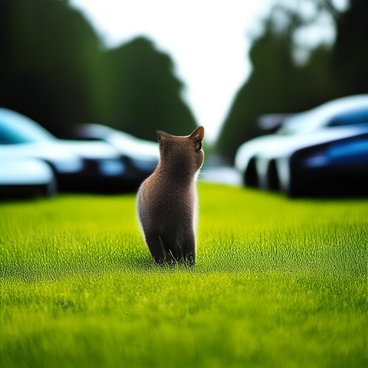}{A cat stands between two parked cars on a grassy sidewalk.}
\visStructured{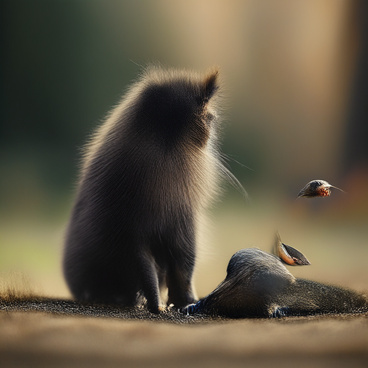}{A long haired cat eating a dead bird.}
\visStructured{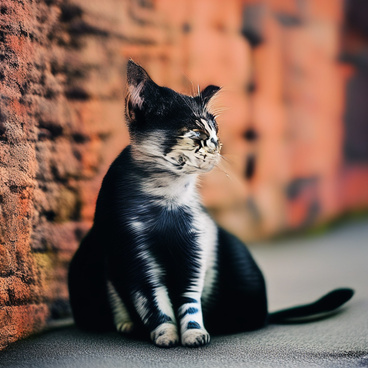}{a stripped cat sitting near a brick wall}
\visStructured{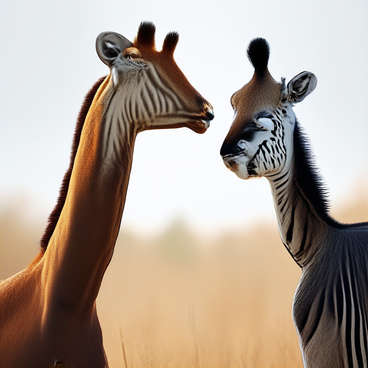}{A giraffe and a zebra checking each other out.}
\visStructured{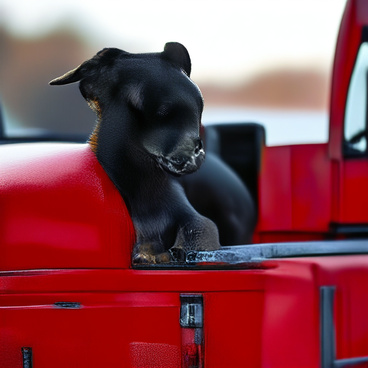}{A red truck has a black dog in the drivers chair.}

\newcommand{\visFlexibility}[2]{
\begin{figure*}[!ht]
    \centering
    \small
    
    % Define image width (4 columns * 0.23 = 0.92\textwidth, leaving space for labels)
    \def\imgw{0.23\textwidth}

    % 1 column for row labels, 4 columns for images
    \begin{tabular}{@{} >{\centering\arraybackslash}m{15pt} c c c c @{}}
        
        % --- Headers ---
        & \textbf{45\% Sparsity} & \textbf{50\% Sparsity} & \textbf{55\% Sparsity} & \textbf{60\% Sparsity} \\[1ex]

        % --- Row 1: OBS-Diff ---
        \rotatebox{90}{\textbf{OBS-Diff}} &
        \adjustbox{valign=m}{\includegraphics[width=\imgw]{figures/Flexibility_PixArt/#1/OBS-Diff_45pct.jpg}} &
        \adjustbox{valign=m}{\includegraphics[width=\imgw]{figures/Flexibility_PixArt/#1/OBS-Diff_50pct.jpg}} &
        \adjustbox{valign=m}{\includegraphics[width=\imgw]{figures/Flexibility_PixArt/#1/OBS-Diff_55pct.jpg}} &
        \adjustbox{valign=m}{\includegraphics[width=\imgw]{figures/Flexibility_PixArt/#1/OBS-Diff_60pct.jpg}} \\[1.5ex]

        % --- Row 2: Ours (CFG) ---
        \rotatebox{90}{\textbf{Ours (CFG)}} &
        \adjustbox{valign=m}{\includegraphics[width=\imgw]{figures/Flexibility_PixArt/#1/Ours_CFG_45pct.jpg}} &
        \adjustbox{valign=m}{\includegraphics[width=\imgw]{figures/Flexibility_PixArt/#1/Ours_CFG_50pct.jpg}} &
        \adjustbox{valign=m}{\includegraphics[width=\imgw]{figures/Flexibility_PixArt/#1/Ours_CFG_55pct.jpg}} &
        \adjustbox{valign=m}{\includegraphics[width=\imgw]{figures/Flexibility_PixArt/#1/Ours_CFG_60pct.jpg}} \\[1.5ex]

        % --- Row 3: Ours (Det.) ---
        \rotatebox{90}{\textbf{Ours (Det.)}} &
        \adjustbox{valign=m}{\includegraphics[width=\imgw]{figures/Flexibility_PixArt/#1/Ours_CFG_Yolo_45pct.jpg}} &
        \adjustbox{valign=m}{\includegraphics[width=\imgw]{figures/Flexibility_PixArt/#1/Ours_CFG_Yolo_50pct.jpg}} &
        \adjustbox{valign=m}{\includegraphics[width=\imgw]{figures/Flexibility_PixArt/#1/Ours_CFG_Yolo_55pct.jpg}} &
        \adjustbox{valign=m}{\includegraphics[width=\imgw]{figures/Flexibility_PixArt/#1/Ours_CFG_Yolo_60pct.jpg}} \\[1.5ex]

        % --- Row 4: Ours (Canny) ---
        \rotatebox{90}{\textbf{Ours (Canny)}} &
        \adjustbox{valign=m}{\includegraphics[width=\imgw]{figures/Flexibility_PixArt/#1/Ours_Canny_45pct.jpg}} &
        \adjustbox{valign=m}{\includegraphics[width=\imgw]{figures/Flexibility_PixArt/#1/Ours_Canny_50pct.jpg}} &
        \adjustbox{valign=m}{\includegraphics[width=\imgw]{figures/Flexibility_PixArt/#1/Ours_Canny_55pct.jpg}} &
        \adjustbox{valign=m}{\includegraphics[width=\imgw]{figures/Flexibility_PixArt/#1/Ours_Canny_60pct.jpg}} \\
        
    \end{tabular}

    \vspace{2mm}
    \caption{\textbf{Different importance signals.}
Comparison between OBS-Diff and our importance-aware variants guided by CFG \cite{ho2022classifier}, CFG+ object-detection\cite{10533619}, and Canny-edge\cite{Canny1986ACA} signals at 45\%, 50\%, 55\%, and 60\% unstructured sparsity. \textbf{Prompt:} "#2"}
    \label{fig:flexibility_#1}
\end{figure*}
}

 % \HL{update all other captions following this one.}
 
\visFlexibility{00000}{A black Honda motorcycle parked in front of a garage.}
\visFlexibility{00620}{A large plant is in the corner of a small bathroom.}
% \visFlexibility{00318}{here}
\visFlexibility{00327}{A dirt bike parked near a tent in the woods.}
\visFlexibility{00509}{A gray and white kitten in a white bathroom sink.}
\visFlexibility{00610}{A large truck is parked beside an old, wooden bench.}

\newpage
\section{Broader Impact}
This work contributes to making text-to-image diffusion models more efficient, practical, and accessible. By reducing model parameters while preserving perceptually important content, importance-aware pruning can lower memory usage, inference cost, and energy consumption. This can broaden access to high-quality generative models beyond large-scale compute environments, enabling more researchers, developers, and practitioners to use diffusion models on limited hardware.

A key positive impact of our work is that it provides a more flexible view of model compression. Instead of treating all generation errors equally, our framework allows pruning to prioritize the visual properties that matter for a given application, such as foreground objects, structural boundaries, or fine image details. This makes compression more adaptable to real deployment needs, where preserving semantically important regions is often more valuable than uniformly minimizing reconstruction error everywhere.

The proposed framework is also model-agnostic and easy to customize. Since the importance signal can come from readily available sources such as CFG responses, edge detectors, or object detectors, the method does not require retraining a separate evaluation model or redesigning the diffusion architecture. This makes it a practical tool for efficient deployment and opens a path toward application-aware compression, where different domains can define their own preservation priorities.

At the same time, importance-aware pruning introduces potential risks. If the importance signal is biased, incomplete, or poorly chosen, the compressed model may systematically preserve some visual concepts while degrading others. Category-targeted pruning can improve fidelity for selected categories, but it could also reduce performance on underrepresented or non-target categories if used carelessly. More generally, controllable compression creates a new axis of model specialization, which should be evaluated not only by average metrics but also by subgroup robustness, object-level fidelity, and failure cases.

This work does not introduce new generative capabilities or modify the training data of the underlying model. Instead, it focuses on improving the efficiency and fidelity of existing models under compression. As a result, its main expected benefits are reduced computational cost, improved deployment feasibility, and better preservation of user-relevant image content after pruning.

More broadly, our results suggest that efficient generative modeling should be evaluated not only by compression ratio, but also by what content is preserved after compression. We hope this encourages future work on controllable, interpretable, and task-aware compression methods for generative models, where efficiency gains are achieved without sacrificing the visual details most important to users.

%%%%%%%%%%%%%%%%%%%%%%%%%%%%%%%%%%%%%%%%%%%%%%%%%%%%%%%%%%%%

\newpage
\FloatBarrier
\end{document}